\documentclass[final,3p,times,twocolumn]{elsarticle}

\usepackage{amssymb}
\usepackage{amsmath}

\usepackage{amsmath,amsfonts,bm}

\def\figref#1{Fig.~\ref{#1}}
\def\Figref#1{Fig.~\ref{#1}}

\def\secref#1{section~\ref{#1}}
\def\Secref#1{Section~\ref{#1}}

\def\eqref#1{Eq.~\ref{#1}}

\def\tabref#1{Table~\ref{#1}}

\def\1{\bm{1}}

\def\rvf{{\mathbf{f}}}

\def\rvs{{\mathbf{s}}}

\def\rvw{{\mathbf{w}}}

\def\rvy{{\mathbf{y}}}

\DeclareMathAlphabet{\mathsfit}{\encodingdefault}{\sfdefault}{m}{sl}
\SetMathAlphabet{\mathsfit}{bold}{\encodingdefault}{\sfdefault}{bx}{n}

\def\sS{{\mathbb{S}}}

\def\sX{{\mathbb{X}}}
\def\sY{{\mathbb{Y}}}

\usepackage{xcolor}
\usepackage{xspace}
\usepackage{float}
\usepackage{wrapfig}
\usepackage{makecell}
\usepackage{multirow}
\usepackage{graphicx}
\usepackage{natbib}

\newcommand{\ie}{i.e.,~}

\newcommand{\eg}{e.g.,~}

\newcommand{\edit}[1]{{#1}}
\newcommand{\revise}[1]{{#1}}

\def\PSM{PSM\xspace}
\def\DAE{DG-AE\xspace}
\def\ISR{ISR\xspace}

\def\DPC{DPC\xspace}
\def\SC{SC3K\xspace}
\def\SCP{SC3K*\xspace}
\def\PS{Point2SSM\xspace}
\def\PSP{Point2SSM++\xspace}

\def\PSMr{PSM \@\cite{cates2017shapeworks}\xspace}

\def\DAEr{DG-AE \@\cite{wang2019dgcnn}\xspace}
\def\ISRr{ISR \@\cite{chen2020isr}\xspace}

\def\DPCr{DPC \@\cite{lang2021dpc}\xspace}

\def\SCr{SC3K \@\cite{zohaib2023sc3k}\xspace}
\def\SCPr{SC3K* \@\cite{zohaib2023sc3k}\xspace}
\def\PSr{Point2SSM \@\cite{adams2023point2ssm}\xspace}

\renewcommand{\S}{\mathbb{S}}

\newcommand{\x}{\mathbf{x}}

\newcommand{\W}{\mathbf{W}}

\newcommand{\F}{\mathbf{F}}

\newcommand{\M}{\mathbf{M}}

\journal{Medical Image Analysis}

\begin{document}

\begin{frontmatter}



\title{Point2SSM++: Self-Supervised Learning of Anatomical Shape Models from Point Clouds}

\author[mymainaddress,mysecondaryaddress]{Jadie Adams\corref{mycorrespondingauthor}}
\cortext[mycorrespondingauthor]{Corresponding author}
\ead{jadie@sci.utah.edu}
\author[mymainaddress,mysecondaryaddress]{Mokshagna Sai Teja Karanam}
\author[mymainaddress,mysecondaryaddress]{Shireen Elhabian}

\address[mymainaddress]{Scientific Computing and Imaging Institute, 72 Central Campus Dr, University of Utah, Salt Lake City, Utah-84112, USA}
\address[mysecondaryaddress]{Kahlert School of Computing, 50 Central Campus Dr, University of Utah, Salt Lake City, Utah-84112, USA}

\begin{abstract}
Correspondence-based statistical shape modeling (SSM) stands as a powerful technology for morphometric analysis in clinical research. SSM facilitates population-level characterization and quantification of anatomical shapes such as bones and organs, aiding in pathology and disease diagnostics and treatment planning. Despite its potential, SSM remains under-utilized in medical research due to the significant overhead associated with automatic construction methods, which demand complete, aligned shape surface representations. Additionally, optimization-based techniques rely on bias-inducing assumptions or templates and have prolonged inference times as the entire cohort is simultaneously optimized. To overcome these challenges, we introduce Point2SSM++, a principled, self-supervised deep learning approach that directly learns correspondence points from point cloud representations of anatomical shapes. Point2SSM++ is robust to misaligned and inconsistent input, providing SSM that accurately samples individual shape surfaces while effectively capturing population-level statistics. Furthermore, we present extensions of Point2SSM++ tailored for dynamic spatiotemporal and multi-anatomy scenarios, showcasing the broad versatility of the framework. Through extensive validation across diverse anatomies, evaluation metrics, and clinically relevant downstream tasks, we demonstrate Point2SSM++'s superiority over existing state-of-the-art deep learning models and the traditional approach. Point2SSM++ substantially enhances the feasibility of SSM generation and significantly broadens its array of potential clinical applications.

\end{abstract}




\end{frontmatter}


\section{Introduction}
\label{sec:intro}
Anatomical shape analysis enables relating form to function and disease, aiding clinical research. 
Statistical shape models (SSM) provide a population-level representation of shapes, such as bones and organs, enabling quantifiable characterization of shape variation and morphometrics.
SSM has numerous critical applications, including pathology characterization \cite{clouthier2023morphable,atkins2017quantitative}, disease staging and diagnosis 
\cite{soufi2019liver, schaufelberger2022cranio},
treatment planning \cite{abler2018statistical,van2023corrective}, and implant and prosthetic design  
\cite{friedrich2023point, sciortino2023statistical}. 
Correspondence-based SSM utilizes an intuitive representation of shape in the form of ordered sets of landmark or correspondence points. 
These points are defined on the surface of shapes in semantically consistent locations across the population. 

Correspondence-based SSM is a powerful tool for quantitative anatomical analysis, but it is underutilized due to the cumbersome construction process. 
Traditional methods automatically define dense sets of correspondence points via optimization schemes.
Such techniques first require segmenting complete, artifact-free representations of shapes from 3D medical images (\ie CT or MRI scans), typically in the form of meshes or binary volumes. This laborious process is done manually or semi-automatically by domain experts. Next, the cohort of shapes must be preprocessed and aligned to factor out global geometric information (i.e., location and orientation). Finally, the correspondence point placement is optimized across the entire cohort at once, typically requiring tedious parameter tuning. The major disadvantages of \revise{optimization-based} SSM construction workflows can be summarized as follows:
\begin{enumerate}
    \item \revise{A strict requirement for complete input shapes (meshes or binary volumes) that are free from noise and artifacts, or the need for additional steps to address such issues.}
    \item A reliance on preprocessing procedures such as shape alignment.
    \item Assumptions embedded in the optimization objectives, which can bias the variability captured by the resulting SSM (\eg restricting variation to linear modes).
    \item Optimization is performed simultaneously on the entire shape cohort, which is time-consuming and hinders real-time and scalable inferences on large cohorts. Optimizing SSM for a new patient requires repeating \revise{steps of the} pipeline.
\end{enumerate}

Progression in the field of point cloud deep learning has enabled reducing the burdens associated with anatomical SSM construction. 
Point clouds provide an unstructured representation of shape that is more readily acquired than complete surfaces such as meshes or binary volumes. 
For example, point clouds can be obtained from lightweight techniques such as thresholding clinical images, anatomical surface scanning, or combining 2D contour representations \cite{timmins2021effect,3Dscanning}.
Since the introduction of PointNet \cite{qi2017pointnet}, point cloud deep networks have been developed for numerous supervised and unsupervised applications \cite{guo2020deep,xiao2023unsupervised}, including tasks closely related to SSM such as keypoint detection and pairwise point matching \cite{chen2020isr,lang2021dpc}.
Inspired by these methods, Point2SSM \cite{adams2023point2ssm} was introduced to infer correspondence-based 
anatomical SSM directly from point clouds in an unsupervised manner. 
\PS demonstrated the potential to learn precise correspondence points in a self-supervised manner. However, it is subject to certain limitations, including the prerequisite of pre-aligning input point clouds and the deficiency of mechanisms to promote consistency across various samplings of the same shape.

In this work, we propose \PSP, a self-supervised approach that addresses the limitations of anatomical SSM construction methods. Our approach significantly extends the preliminary \PS model introduced in \cite{adams2023point2ssm}, offering a principled theoretical framework for how correspondence is promoted, addressing its limitations, and expanding its potential clinical application. Our contributions are summarized as follows:
\begin{itemize}
    \item We propose \PSP, which improves upon \PS in two important ways. First, we adapt the network to allow for misaligned input point clouds, removing the need for any preprocessing steps. Second, we modify the loss to encourage sampling invariance and rotation equivariance in output correspondence points. 
    \item We provide a principled discussion of the inductive biases in \PS and \PSP that promote point correspondence across shape samples.
    \item  We conduct thorough experimentation on multiple datasets with comprehensive evaluation metrics and benchmarking against state-of-the-art (SOTA) models. Furthermore, we assess performance on two downstream clinical tasks, showcasing practical usability.
    \item We adapt \PSP to allow for multi-anatomy learning so that a single model can be used to infer SSM of multiple anatomical shapes. 
    \item We showcase how \PSP can be adapted to capture 4D, spatiotemporal data, enabling longitudinal clinical studies and dynamic shape analysis.
\end{itemize}

The paper is organized as follows. \Secref{sec:literature} provides a review of existing literature. 
\Secref{sec:methods} details our proposed model, \PSP, its architecture, training procedures, and theoretical underpinnings. Sections \ref{sec:nonlinear} through \ref{sec:4d} provide thorough evaluations of \PSP. 
Finally, \Secref{sec:conclusion} summarizes our findings and future research. 
\section{Related Work}
\label{sec:literature}

Historically, correspondence points for shape analysis were defined manually to capture biologically relevant features \cite{dryden2016statistical}. 
Computational approaches later automated dense correspondence construction via metrics such as entropy \cite{cates2007shape} or minimum description length \cite{davies2002minimum}, or via parametric representations \cite{ovsjanikov2012functional,styner2006framework}.
Correspondence-based SSM aims to provide both accurate surface sampling (via well-distributed points constrained to the surface) and effective representation of the population shape statistics (via good correspondence). 
Particle-based shape modeling (PSM) \cite{cates2007shape,cates2017shapeworks} is the SOTA optimization-based technique for SSM construction, achieving both of these goals \cite{goparaju2022benchmarking}.
\PSM has additionally been adapted to capture temporal trajectories in dynamic or longitudinal data (\ie sequences of shape over time) \cite{datar2009regression,adams2022disentangled4d,adams2022polynomial4d}.

Nevertheless, optimization-based approaches such as \PSM have practical challenges, as they often depend on complete, high-quality input shapes, and inference can be computationally intensive. Moreover, the entropy-based optimization scheme utilized by \PSM assumes primarily linear correlations, which can bias the representation of population variation.

Deep learning methods address these limitations by predicting SSMs directly.
Supervised models learn correspondences from images \cite{bhalodia2024deepssm,uncertaindeepssm,adams2022images,adams2023fully,ukey2023localization}, reducing inference time but depending on optimization-based correspondences for training.
Other approaches have sought to reduce limitations in both training and testing via unsupervised estimation of SSM \cite{xu2023image2ssm,iyer2023mesh2ssm,adams2023point2ssm,adams2023point}.
Mesh2SSM \cite{iyer2023mesh2ssm} predicts correspondence-based SSM in an unsupervised manner from complete mesh shape representations. 
Point2SSM \cite{adams2023point2ssm} is the only model specifically developed to predict anatomical SSM from raw point clouds.


\edit{Beyond correspondence-based approaches, several recent methods leverage continuous neural representations to capture population-level shape variation.
ShapeFlow \cite{jiang2020shapeflow} introduced learnable deformation flows for shape interpolation and correspondence.
Building on this idea, landmark-free group-wise SSM construction has been achieved using neural deformation fields \cite{ludke2022landmark}, which model smooth transformations across populations.
More recently, \citet{amiranashvili2024learning} proposed to learn continuous shape priors from sparse data using neural implicit functions, enabling SSM construction without requiring dense input representations.
Together, these works highlight a growing trend toward continuous, data-driven models of shape populations that bypass traditional correspondence optimization.
These deformation-based approaches assume the availability of a template shape and a surface or volumetric parameterization, from which correspondences can be propagated via the learned diffeomorphic mappings. In contrast, \PSP is designed for settings where only unordered point clouds are available and no template surface, segmentation, or predefined parameterization exists. While diffeomorphic models offer the flexibility to transfer arbitrary expert-defined landmarks through the deformation field, they operate under fundamentally different assumptions than point-cloud–only methods. We therefore view deformation-based SSMs and correspondence-learning approaches such as \PSP as complementary: the former are well suited to template-based shape analysis pipelines, whereas the latter target data-scarce or minimally processed point cloud scenarios where explicit template construction is impractical.}

In parallel, deep learning directly on point clouds has become an active research area.
PointNet \cite{qi2017pointnet} was the pioneering deep network designed to process point cloud input in a permutation-invariant manner via independently learning local features for each point, then aggregating to acquire a global feature vector. PointNet++ \cite{qi2017pointnet++} built upon this architecture by incorporating a hierarchical structure to learn multiscale geometric information.
This independent treatment of points neglects the geometric relationships among points, inspiring the dynamic graph convolution neural net (DGCNN) \cite{wang2019dgcnn}. DGCNN defined edge convolution (EdgeConv), which captures local geometric structure while maintaining permutation invariance. Point cloud convolution has also been expanded to 4D point cloud sequences \cite{fan2021deep,fan2022pstnet}. 
Self-attention methods offer a flexible way to capture complex relationships among point features, enabling more expressive point cloud representations \cite{sun2020pointgrow,wang2022pointattn}.

Various unsupervised methods in computer vision have been developed to establish shape correspondence, either from a pairwise or class-level global perspective. 
Pairwise approaches aim to map points from a source shape to a target shape, either by learning deformations to a predefined template \cite{deprelle2019learning,cosmo2016matching} or matching permutations \cite{chen2020isr,lang2021dpc}. 
\citet{zeng2021corrnet3d} employs an encoder-decoder architecture for shape coordinate regression, while \citet{lang2021dpc} skip the decoder and reconstructs using the original point cloud and similarity learned in the learned feature space.
A rotation-invariant transformer for point cloud matching was proposed in \cite{yu2023rotation}, which improves point cloud matching by considering rotational invariance.
Similarly, \citet{lv2023kss} uses Kendall shape space for point cloud registration, providing more robust matching.
Global correspondence methods have focused on identifying class-specific keypoints from point clouds \cite{suwajanakorn2018keypointnet, fernandez2020unsupervised,jakab2021keypointdeformer, cheng2021learning,chen2020isr}. 
SC3K is a recently proposed model that accurately infers sparse key points given imperfect input point clouds but struggles to predict dense output points \cite{zohaib2023sc3k}.
These approaches establish correspondence but were not developed for the task of anatomical SSM,  
where the training sample size is typically limited, and dense points are required to capture anatomical detail. 
\citet{adams2023point} found some success in applying point completion networks to the task of anatomical SSM. 
In encoder-decoder-based point completion networks, a learned low-dimensional latent representation is mapped continuously to the output space. This bottleneck results in output correspondence. Inspired by this potential, Point2SSM \cite{adams2023point2ssm} was proposed to address the specific challenges of anatomical SSM, including robustness to data scarcity and imperfect input (\ie noisy, partial, or sparse point clouds). Point2SSM learns output correspondence points as a weighted combination of the input point cloud in a permutation-invariant and self-supervised manner, making it robust to limited training data and large shape variation. 

\section{Methods}
\label{sec:methods}
This section provides an overview of Point2SSM and the enhancements provided by Point2SSM++, including a principled explanation of how correspondence is achieved, adaptations for 4D and multi-anatomy modeling, and evaluation metrics.
\edit{In this work, we use the term shape model to refer to a model that predicts anatomically consistent correspondence points across a population. Point2SSM and Point2SSM++ are therefore correspondence-learning models that produce a common parameterization of shape instances. They do not, by themselves, impose a probabilistic or linear model of shape variation (e.g., PCA) in the resulting shape space. Instead, the learned correspondences may be used as input to downstream statistical shape modeling techniques, including linear or nonlinear SSMs, as is standard practice in correspondence-based shape analysis.}

\subsection{Notation}

Let $\sS$ denote a shape represented by a point cloud, meaning an unordered set of $I$ points. Then  $\sS = \{\rvs_1, \dots, \rvs_I \}$ where $\rvs_i \in \mathbb{R}^3$ is a 3D vector containing the $x$, $y$, and $z$ coordinates of the $i$th point. 
The input of Point2SSM is a subset of $N$ points in $\sS$ (where $N < I$), denoted $\sX$: $\sX = \{\x_1, \dots, \x_N \}$ where $\x_n \in \sS$. 
The output of Point2SSM is a set of $M$ correspondence points, denoted $\sY$: $\sY = \{\rvy_1, \dots, \rvy_M \}$ where $\rvy_m \in \mathbb{R}^3$. There are no restrictions on the value of $M$, but Point2SSM achieves the best performance when $M \leq N$ \cite{adams2023point2ssm}.
Given two different shapes of the same anatomy type, $\sS^{(1)}$ and $\sS^{(2)}$, we expect outputs $\sY^{(1)}$ and $\sY^{(2)}$ to be in correspondence, meaning point $\rvy^{(1)}_m$ is in a semantically consistent location to point $\rvy^{(2)}_m$ for all $m \in \{1, \dots, M\}$.

\subsection{Point2SSM}
\label{sec:Point2SSM}

Point2SSM defines the output as the Hadamard product of the input point cloud and a learned attention map, $\W \in \mathbb{R}^{M \times N}$ with elements $w_{m,n}$.
Specifically, Point2SSM estimates each output correspondence point, $\rvy_m$, as a weighted combination of all input points in $\sX$:
\begin{equation}
    \rvy_m = \sum_{n=1}^N w_{m,n}*\x_n
    \label{eq:ym}
\end{equation}
As a result, the output points are constrained to be within the convex hull of the input point cloud. This constraint accelerates learning and improves surface sampling accuracy by requiring predicted points to be close to the shape surface. 
\subsubsection{Architecture}

\begin{figure*}[ht!]
    \centering
    \includegraphics[width=.72\textwidth]{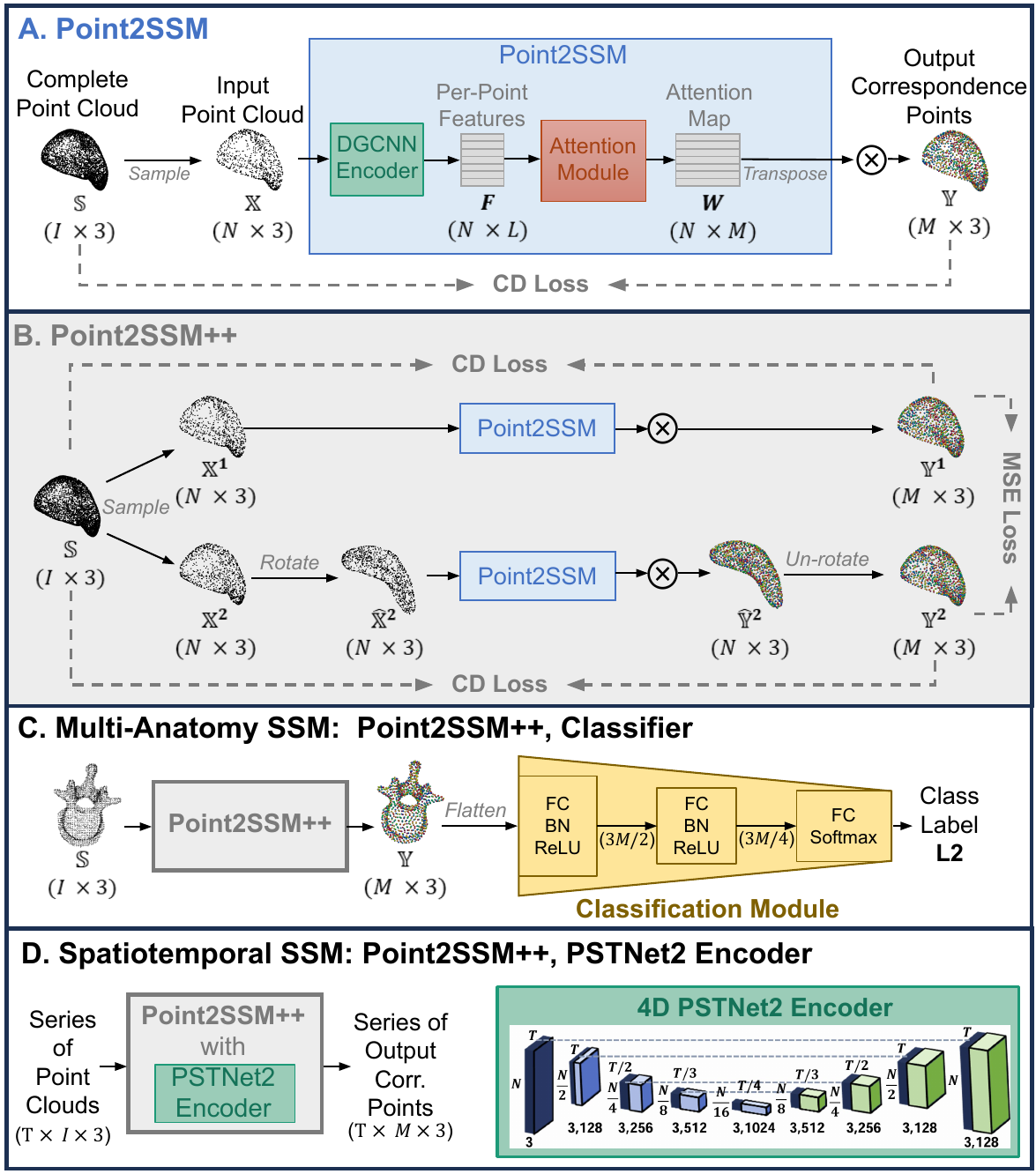}
    \caption{A. Point2SSM \cite{adams2023point2ssm} architecture B. Point2SSM++ architecture C. Point2SSM++, Classifer architecture for multi-anatomy SSM. The classification modeule is a simple multilayer perceptron comprised of fully connected (FC) layers, batch normalization (BN) and activation.  D. Point2SSM++ with 4D PSTNet2 \cite{fan2021pstnet2} encoder for spatiotemporal SSM given sequences of anatomical shapes over time.}
    \label{fig:arch}
\end{figure*}

As shown in Fig. \ref{fig:arch}.A, the Point2SSM model consists of two subnetworks: the \textbf{encoder}, which learns a feature representation of each input point and its neighbors, and the \textbf{attention module}, which estimates an attention map $\W$ from the features. 
DGCNN \cite{wang2019dgcnn} is selected for the encoder architecture, enabling topological information to be learned via edge convolution. 

The encoder takes a point $\x_n \in \sX$ and its $K$ nearest neighbors (with respect to Euclidean distance) as input and produces an $L$-dimensional feature representation $\rvf_n \in \mathbb{R}^L$ (corresponding to the $n$-th row in the learned feature matrix $\F$). By incorporating the spatial relation between points, the DGCNN encoder provides features that capture the local semantic characteristics of the shape.
The attention module takes a feature vector $\rvf_n$ as input and outputs $\rvw_n \in \mathbb{R}^M$ (representing the $n$-th row in the weight map $\W$). 
The vector $\rvw_n$ captures the degree to which input point $n$ contributes to determining the positions of each of the $M$ output correspondence points (see \eqref{eq:ym}).
Point2SSM and its extension are \textbf{permutation invariant} because each row of the attention map is independently determined from a given point and its neighbors, \ie permuting the input $\sX$ identically permutes the rows of $\W$ providing the same $\sY$.
The attention module utilizes self-attention to predict the attention map from the feature representation via self-feature augment (SFA) blocks originally introduced in \cite{wang2022pointattn}. 
The SFA block is a multihead, self-attention mechanism that draws on the core idea of transformers to optimally integrate the information from different point features.

\subsubsection{Loss}
In Point2SSM, the attention map, $\W$, is learned in a self-supervised manner by encouraging $\sY$ to match the full point clouds $\sS$.
The Point2SSM loss comprises two terms - one that encourages $\sY$ to reconstruct $\sS$ and one that regularizes $\sY$ across a minibatch, encouraging better correspondence. 
Chamfer distance (CD) is used for the reconstruction loss term, as it quantifies point cloud differences in a permutation-invariant way:
\begin{equation}
    \mathrm{CD}(\sY, \sS) = 
    \frac{1}{|\sY|} \sum_{\rvy\in\sY} \min_{\rvs \in \sS} || \rvy - \rvs ||_2^2 + \frac{1}{|\sS|} \sum_{\rvs\in\sS}\min_{\rvy\in\sY} || \rvs - \rvy||_2^2
    \label{eq:CD}
\end{equation} 

Other point cloud distance metrics have been proposed, such as Cauchy-Schwartz divergence \cite{he2023learning} and variants of Earth Mover's distance \cite{nguyen2021swd}. However, as supplementary experiments in \ref{app:distance} show, Chamfer Distance performs best on this task.
Note that while Point2SSM is provided only with a subset of $\sS$ as input, denoted $\sX$, the CD loss is computed between $\sY$ and the full $\sS$.
In training, $\sX$ is randomly sampled from $\sS$ in each iteration, serving as an automatic form of data augmentation and leveraging the benefits of masked modeling. This approach improves model robustness as demonstrated by supplementary experiments in \ref{app:distance}. 

A pairwise mapping error (ME) \cite{lang2021dpc} provides a regularization term to improve correspondence. The ME between two output point clouds $\sY^{(1)}$ and $\sY^{(2)}$ is defined as:
\begin{equation}
    \text{ME}(\sY^{(1)}, \sY^{(2)})=\frac{1}{M*R} \sum_{m=1}^M \sum_{r\in \mathcal{N}_{(1)}\left(\rvy^{(1)}_m\right)} v^{(1)}_{m r}\left\|\rvy^{(2)}_m-\rvy^{(2)}_r\right\|_2^2
    \label{eq:ME}
\end{equation} 
where $\mathcal{N}_{(1)}\left(\rvy^{(1)}_m\right)$ are the $R$-indices of the $R$-nearest neighbors of point $\rvy^{(1)}_m$ in $\sY^{(1)}$ (in Euclidean distance). 
The term $v^{(1)}_{m r}=e^{-\left\|\rvy^{(1)}_m-\rvy^{(1)}_r\right\|_2^2}$ weights the loss elements according to the proximity of the neighbor points. 
The ME loss encourages point neighborhoods in $\sY^{(1)}$ to match those in $\sY^{(2)}$. Consistent point neighborhoods are an indication of accurate correspondence across shapes.  ME is computed between all pairs of output in a minibatch of size $B$. 

Given a batch of point cloud shape representations denoted $\mathcal{S}_B$ and predictions $\mathcal{Y}_B$, the Point2SSM loss is defined as:
 \begin{align}
      \mathcal{L}&_{\text{Point2SSM}}(\mathcal{S}_B, \mathcal{Y}_B) = \frac{1}{B}\sum_{i=1}^B\mathrm{CD}(\sY^{(i)}, \sS^{(i)}) + \nonumber \\
      &\alpha \left(\frac{1}{(B-1)^2}\sum_{i=1}^B \sum_{j=1, j\neq i}^B \mathrm{ME}(\sY^{(i)},\sY^{(j)}) + \mathrm{ME}(\sY^{(j)},\sY^{(i)}) \right) \nonumber \\ 
      \label{eq:loss}
 \end{align}
 where $\alpha$ is a hyperparameter that controls the effect of the regularization \cite{adams2023point2ssm}. In this work, $\alpha$ is set to 0.1 based on tuning using the validation set. \PS accuracy is not particularly sensitive to the choice of $\alpha$ \cite{adams2023point2ssm}. 

\subsubsection{Inductive Biases that Promote Learning Correspondence}

\edit{Point2SSM has been empirically shown to produce correspondence points that are consistent across a population, despite the absence of explicit correspondence supervision. This behavior arises from inductive biases in the architecture and loss design rather than from an explicit statistical model of shape variation.}
This effect can be understood from the lens of information bottleneck (IB) theory, where a latent encoding is learned to capture the minimal sufficient statistics required of the input to predict the output \cite{tishbyIB}.
The CD loss (\eqref{eq:CD}) indirectly requires the features $\F$ to capture significant information from $\sX$ relevant for accurately predicting $\sY$.
The parameter $K$, determining the size of the neighborhood considered for each point, influences the degree of local versus global information captured in each feature encoding $\rvf_n$. Hence, the choice of $K$ controls the granularity of the information extracted for each point in the point cloud.
This compression aligns with the IB principle's aim to minimize the entropy between the input and latent encoding. The parameter $K$ must be optimally selected to capture enough local information without overfitting or including irrelevant information, as demonstrated by supplementary experiments in 
\ref{app:k_ablation}.
The attention module assigns weights to the feature vectors based on their relevance for generating $\sY$, effectively allowing the network to focus on the most informative aspects of $\F$ for predicting $\sY$. This process aligns with the IB principle's aim to maximize the entropy between the latent encoding and output. 
In this manner, Point2SSM learns to discard redundant or noisy information in $\sX$ that does not contribute to the accurate prediction of $\sY$, thereby enhancing the network's ability to generalize and identify correspondences based on robust and distinctive features.
At the same time, the attention module ensures that the retained information in $\F$ is maximally useful for the reconstruction task, which entails identifying and leveraging patterns (correspondences) that are consistent across different instances of the shape.

\subsection{Point2SSM++}
\label{sec:Point2SSM++}

This section details the proposed approach, \PSP. \PSP improves output consistency by augmenting two different inputs $\sX$ and $\widetilde{\sX}$, of the same point cloud $\sS$ as shown in \figref{fig:arch}.B. \PSP also allows for misaligned input and can be adapted for multi-anatomy (\figref{fig:arch}.C) and spatiotemporal scenarios (\figref{fig:arch}.D), as explained in the following subsections. 

\subsubsection{Allowing for Misaligned Input}
Point2SSM assumes that input shapes are prealigned to remove the impact of translation, rotation, and scale in output predictions. Point2SSM++ removes this requirement, allowing for accurate predictions given input in any coordinate frame. 
This is partially accomplished by automatically normalizing input point clouds to be zero-centered with a uniform scale. 
Centering is performed by translating each point cloud center of mass to the origin and scaling is applied so that the maximum distance of any point to the origin is one.
The inverse of these transforms is then applied to network output to provide predictions in the original coordinate system. 
Rotation generally cannot be automatically factored out as there is no notion of standard orientation. However, point clouds acquired from medical imaging data have an innate orientation resulting from the scanning process. Point2SSM++ capitalizes on this semantic orientation but accounts for small differences in patient pose via a consistency loss, as explained in the following section.
In this manner, Point2SSM++ provides consistent predictions in the original input coordinate systems.  In SSM analysis, correspondence points are mapped to a global coordinate system via generalized Procrustes \cite{gower1975procrustes} so that translation, rotation, and, optionally, scale are not captured as modes of variation.  
An experiment qualitatively demonstrating the efficacy of this approach for handling misaligned input is provided in \ref{app:alignment}.

\subsubsection{Point2SSM++ Consistency Loss}

Accurate SSM construction methods should provide consistent correspondence points given different samplings of the same input shape. As is illustrated in \figref{fig:arch}.B, Point2SSM++ directly encourages this quality by quantifying the consistency in predictions resulting from two different random subsamples, $\sX$ and $\widetilde{\sX}$, of the same point cloud $\sS$.
Additionally, accurate correspondence points should be unaffected by pose, i.e., rotationally equivariant. 
To account for pose, Point2SSM++ samples random $x$, $y$, and $z$ rotations within $\pm 15$ degrees and applies them to input $\widetilde{\sX}$. The inverse rotation is then applied to the resulting prediction to get $\widetilde{\sY}$. The consistency between the resulting predictions, $\sY$ and $\widetilde{\sY}$ is then quantified via mean square error (MSE). Note rotation augmentation could be added to $\sX$ in addition to $\widetilde{\sX}$; however, we elect to preserve the original orientation since it retains semantic consistency in clinical scenarios.

The Point2SSM++ loss is formulated as the Point2SSM loss (\eqref{eq:loss}) computed on both $\sY$ and $\widetilde{\sY}$, as well as this consistency term. Thus, given a batch of point clouds $\mathcal{S}_B$ and predicted output $\mathcal{Y}_B$, the \PSP loss is defined as:

\begin{align}
      \mathcal{L}_{\text{Point2SSM++}}(\mathcal{S}_B, \mathcal{Y}_B) &= \mathcal{L}_{\text{Point2SSM}}(\mathcal{Y}_B,\mathcal{S}_B) \nonumber \\ 
      &+  \mathcal{L}_{\text{Point2SSM}}(\widetilde{\mathcal{Y}}_B,\mathcal{S}_B)\nonumber \\
      &+ c*\frac{1}{B}\sum_{i=1}^B\mathrm{MSE}(\sY^{(i)}, \sY^{(i)*})
      \label{eq:consist_loss}
\end{align}

where $c$ is a scalar hyperparameter, determining the weight of the consistency term. In this work, $c$ is set to 1 in all experiments based on tuning using the validation set.
This loss encourages surface sampling invariance and rotation equivariance, making Point2SSM robust to differences in pose and improving correspondence accuracy.

\subsubsection{Point2SSM++ Multi-Anatomy Training}
\label{sec:multianatomy_methods}
Training \PSr on multiple anatomies was shown to provide similar accuracy to training individual single-anatomy models. 
\PSP can be trained across multiple anatomies, allowing it to leverage shared shape information to improve correspondence predictions. This is particularly beneficial when training data for a specific anatomy type is limited. The training and inference procedures remain the same as in the single-anatomy setting. For evaluation, Procrustes alignment and metric calculation are performed separately for each anatomy to capture anatomy-specific statistics and enable comparison with single-anatomy models.

A concern in multi-anatomy training is the potential loss of anatomy-specific details and features. To mitigate this effect, we propose augmenting the \PSP model with an auxiliary anatomy classifier, depicted in \figref{fig:arch}.C. The classification module is a simple multi-layer perceptron (MLP) comprised of fully connected layers with batch normalization, trained to predict the anatomy class from $\sY$. Its weighted negative log-likelihood (NLL) classification loss term is added to the main loss in Eq. \ref{eq:loss}, encouraging the model to preserve the anatomy-specific morphological characteristics in $\sY$ that are relevant to classification. As a result, the multi-anatomy SSM achieves cross-anatomy correspondence while maintaining distinct features between anatomies.

\subsubsection{Spatiotemporal (4D) Point2SSM++}
\label{sec:4d_methods}
Many clinical research studies require assessing dynamic or longitudinal cases of 4D data or sequences of shapes over time. 
In order to extend the \PSP method to 4D data, we propose replacing the 3D DGCNN encoder with a state-of-the-art point cloud sequence convolutional network: the hierarchical Point Spatio-Temporal net (PSTnet2) \cite{fan2021pstnet2} ( \figref{fig:arch}.D). 
In the \PSP model with PSTNet2 encoder, denoted \textbf{4D PSTNet2 \PSP}, the loss and attention module remain unchanged and are applied cross-sectionally. The key difference lies in the encoder: PSTNet2 captures temporal dependencies and spatial variations jointly, enabling the network to learn rich 4D point cloud representations. This allows the model to incorporate temporal shape dynamics into correspondence prediction, which is crucial for studying anatomical changes over time in clinical applications.

\subsection{Evaluation Metrics}
\label{sec:metrics}

The quality of correspondence-based SSM relies on two key factors: how well points are constrained to and capture the shape surfaces and how well the population-level statistics are captured through correspondence. These aspects are not mutually dependent; as a strong correspondence model might inadequately sample underlying surfaces, and vice versa. 
We utilize nine complementary metrics for thorough evaluation:
    \subsubsection{Surface Sampling}
        \begin{itemize}
            \item \textbf{Chamfer distance (CD)}: A small CD (\eqref{eq:CD}) indicates the output points accurately capture the complete shape. 
            \item \textbf{Point-to-surface distance (P2S)}: P2S quantifies the distance of the output points to a complete surface shape representation (\ie mesh). Low P2S indicates that predicted points are well constrained to the shape surface.
        \end{itemize}
    \subsubsection{Consistency}
        \begin{itemize}
            \item \textbf{Sampling invariance (Samp. MSE)}: Accurate predictions should be invariant to different input samplings of the same shape. This metric is evaluated as the mean square error (MSE) between predictions made from two different random subsamples of the same test point cloud.
            \item \textbf{Rotation equivariance (Rot. MSE)}: Accurate predictions should be equivariant to differences in pose or rotation. This metric is evaluated as the mean square error (MSE) between aligned predictions made from two different random rotations of the same test input.
        \end{itemize}
    \subsubsection{Correspondence}
        \begin{itemize}
            \item \textbf{Mapping error (ME)}: The ME (\eqref{eq:ME}) quantifies how consistent output point neighborhoods are across the population. A lower ME indicates consistent neighborhoods, implying better correspondence.
            \item \textbf{Warp-based reconstruction surface-to-surface distance (Warp S2S)}: Accurate correspondence points can serve as control points for surface mesh warping. Given two sets of correspondence points $\sY^{(1)}$ and $\sY^{(2)}$ and corresponding surface mesh 
            $\M^{(1)}$, a reconstructed surface mesh $\widehat{\M}^{(1)}$ can be obtained by applying the warp between $\sY^{(2)}$ and $\sY^{(1)}$ to $\M^{(2)}$. A smaller surface-to-surface distance between $\M^{(1)}$ and $\widehat{\M}^{(1)}$ indicates better correspondence between the control points.
        \end{itemize}
    \subsubsection{SSM Metrics} Principal component analysis (PCA) is used in SSM analysis to quantify the modes of variation in the population and assess how well population-level statistics are captured \cite{munsell2008eval}. There are three metrics to this end: 
        \begin{itemize}
            \item \textbf{Compactness (Comp.)}: Accurate correspondence leads to a more compact SSM, meaning the training data distribution is represented using the minimum number of parameters. When there is strong correspondence, a large proportion of explained population variance is captured in fewer PCA modes. Thus, a larger area under the cumulative variance plot indicates better correspondence. 
            \item \textbf{Generalization (Gen. CD)}: Accurate SSM should generalize well from training examples to unseen examples. 
            The generalization metric quantifies the CD between the estimated correspondences from test point clouds and their reconstructions from the training SSM-based PCA embeddings with various numbers of components. 
            A smaller CD indicates that the SSM generalizes better. 
            \item \textbf{Specificity (Spec. CD)}:  Specificity measures whether the SSM generates valid instances of the shape class. It is computed as the average CD between training examples and generated samples from the training SSM-based PCA embeddings with various numbers of components. A smaller CD indicates the SSM is more specific.
        \end{itemize}
\section{Experiments}

In all experiments, unless otherwise specified: $|\sS| \approx 5000$, $|\sX| =N = 1024$, $|\sY| = M = 1024$, $K = 26$, $L=128$, and batch size $B = 8$ or $B = 4$ in the case of Point2SSM++.
These values were selected to be consistent with existing work. Point2SSM has been shown to be robust to the choice of $N$, $M$, $B$ \cite{adams2023point2ssm}, and $K$ (\ref{app:k_ablation}).
Adam optimization with a constant learning rate of 0.0001 is used on a 4x TITAN V GPU to train models. \PSP training and inference can be run on a single GPU with high-resolution point clouds. Model training is run until convergence (\ie when CD on the validation set has not improved in 100 epochs). All evaluations utilize the models resulting from the epoch with the best CD validation.

\subsection{Nonlinear Variation Proof of Concept}
\label{sec:nonlinear} 

\begin{figure}[!ht]
    \centering
    \includegraphics[width=.3\textwidth]{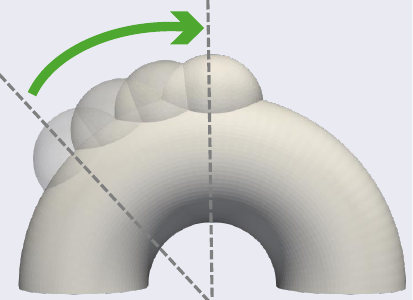}
  \caption{Mode of variation in half torus bump cohort.}
  \label{fig:htb}
\end{figure}

Point cloud deep learning-based SSM generation techniques, such as Point2SSM++, are data-driven and do not rely on any limiting assumptions about the underlying data distribution. Traditional SSM generation methods require assumptions to define optimization objectives or require a template or atlas. 
\PSM \cite{cates2007shape} utilizes a Gaussian entropy-based optimization objective that imposes a linearity assumption, biasing the population variation captured.
To illustrate this difference, we utilize a synthetic 
dataset with known nonlinear shape variation,
comprising half torus shapes with a protruding bump. The shapes are identical across the cohort with the exception of the position of the bump, which varies along a curve, as illustrated in Figure \ref{fig:htb}. Thus, this dataset exhibits a singular, nonlinear mode of variation.\\

\begin{figure}[!ht]
    \centering
    \includegraphics[width=.48\textwidth]{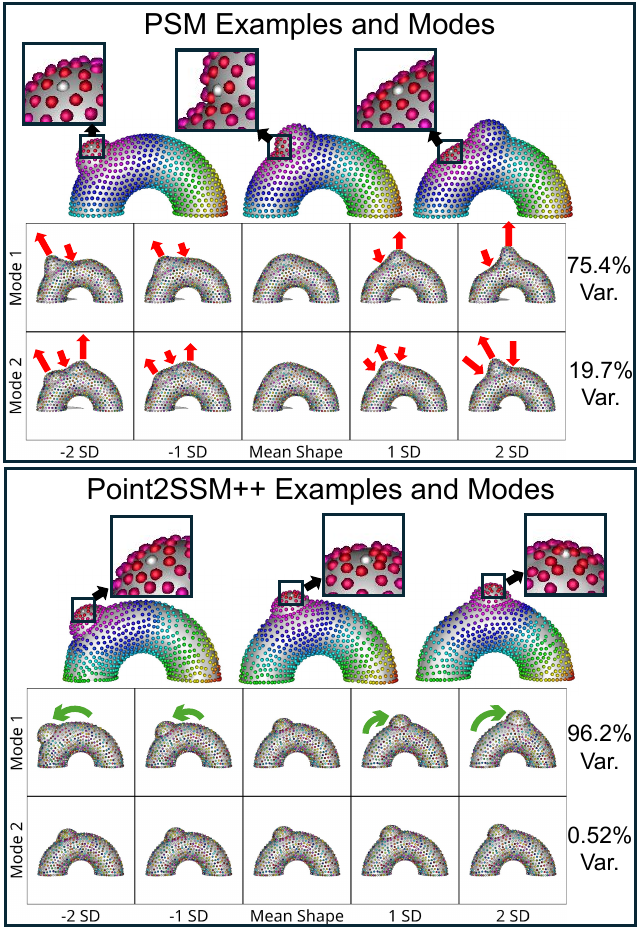}
    \caption{Results on the nonlinear half torus bump dataset from \PSM and \PSP. The three example test predictions (top) and resulting primary and secondary modes of variation (bottom) demonstrate that \PSP correctly captures nonlinear variation, whereas \PSM fails to do so.}
    \label{fig:htb_modes}
\end{figure}

\Figref{fig:htb_modes} provides example correspondence points and the primary and secondary modes of variation resulting from \PSM and \PSP. A single point on the bump is highlighted across three test predictions, with color denoting distance to the point. In the \PSP model, this point consistently stays on the bump across the population as expected, whereas the \PSM model fails to maintain this correspondence. This result is further illustrated by the modes of variation, where \PSP correctly finds the singular true nonlinear mode, whereas \PSM fabricates two linear modes to account for the variation.  Furthermore, the mean shape resulting from our approach resembles a valid instance of the shape class, whereas the mean shape from PSM does not. 

\subsection{Single Anatomy Experiments}
\label{sec:benchmark}
We utilize challenging single anatomy datasets to comprehensively analyze the performance of \PSP compared to \PSM and point cloud deep learning-based methods. The following subsections outline the datasets and comparison models and provide results and downstream task analysis.

\subsubsection{Datasets}
\label{sec:datasets}

We employ publicly available segmentation datasets comprising six distinct anatomies with varying cohort sizes. 
These datasets feature complete mesh shape representations, facilitating comparison with the traditional \PSM method. The unordered mesh vertices serve as the ground truth complete point clouds. All datasets undergo random partitioning into training, validation, and testing sets using an $80\%/10\%/10\%$ split. Each dataset is described as follows:
\begin{itemize}
    \item \textbf{Spleen (40 shapes)}\cite{simpson2019medseg}: The spleen organ dataset provides a limited data scenario with challenging shapes that vary greatly in size and curvature.
    \item \textbf{Pancreas (272 shapes)}\cite{simpson2019medseg}: The pancreas dataset comprises of pancreas organs with tumors of varying sizes from cancer patients, providing difficult patient-specific shape variability. 
    \item \textbf{Liver (834 shapes)} \cite{Ma-2021-AbdomenCT-1K}: The liver dataset provides a organ datasets with nonlinear shape variation.
    \item \textbf{Femur (56 shapes)}: The femur dataset contains proximal femur bones clipped under the lesser trochanter to focus on the femoral head. Nine of the femurs have the cam-FAI pathology characterized by an abnormal bone growth lesion that causes hip osteoarthritis \cite{atkins2017quantitative}.
    \item \textbf{L4 Vert. (160 shapes)}\cite{sekuboyina2021verse}: This dataset contains fourth lumbar (L4) vertebrae bones, with complex topology. 
    \item \textbf{Cranium (400 shapes)}\cite{schaufelberger2022cranio}: This dataset contains 3D head models constructed from surface scans of craniosynostosis patients. There are four balanced classes within this dataset: control and three pathological suture fusion classes (coronal, metopic, and sagittal).
\end{itemize}

\begin{figure*}[!h]
    \centering
    \includegraphics[width=\textwidth]{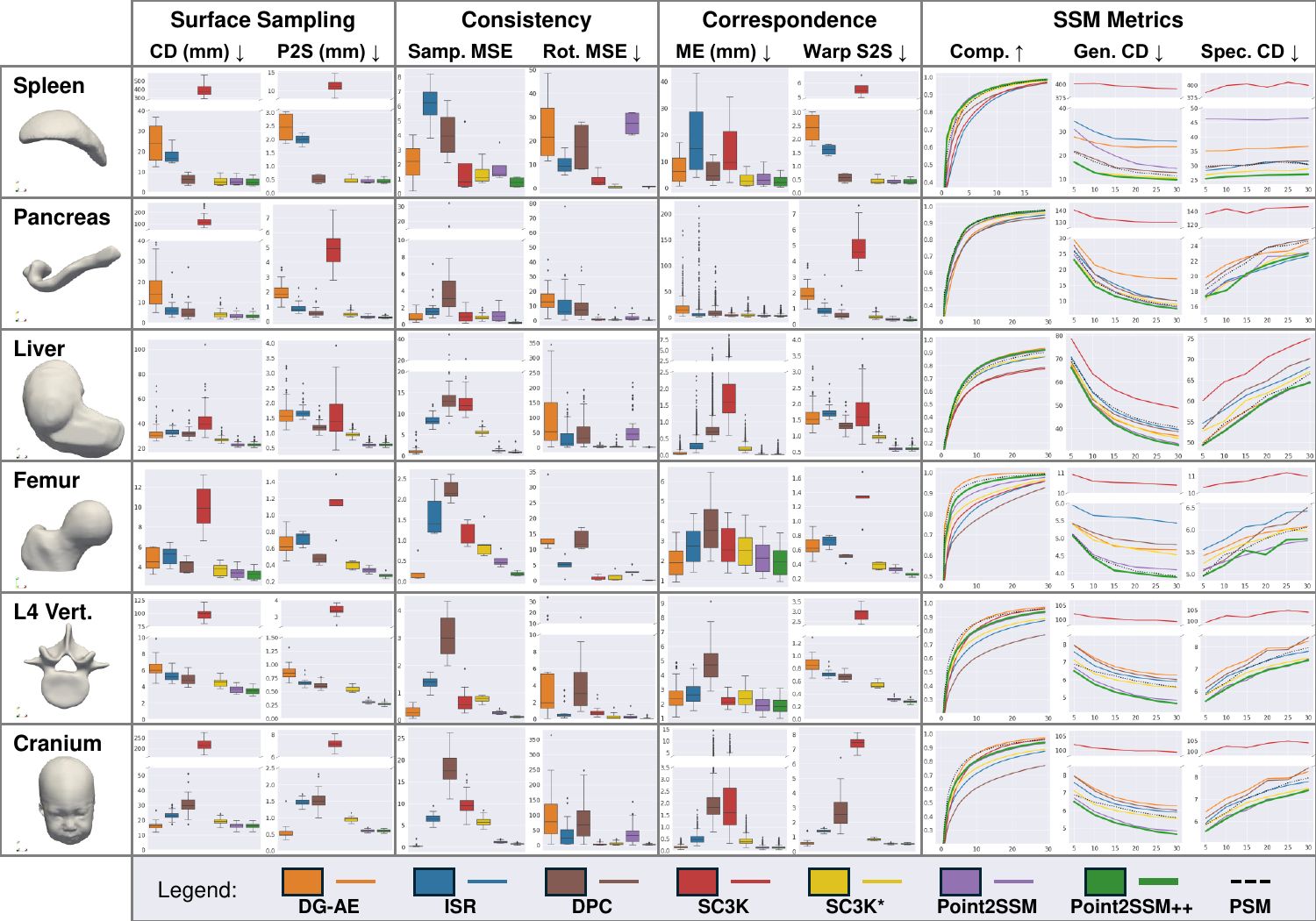}%
    \caption{Single anatomy experiment results. Boxplots show the error distribution across test sets for each model in mm \revise{(see \ref{app:single_pvalues} for analysis of statistical significance).} Compactness plots show the cumulative variance ratio over the number of PCA modes. Generalization and specificity plots show the CD over the number of PCA modes. A maximum of 30 modes is displayed which captures at least $99\%$ of the total variation for all datasets. Lower values are better for all metrics except PCA compactness.}
    \label{fig:benchmark}
\end{figure*}

\subsubsection{Comparison Models}
\label{sec:models}

We compare \textbf{\PS} and \textbf{\PSP} with the traditional \PSM model, as well as SOTA point cloud deep learning based models. Each method is described as follows:
\begin{itemize}
    \item \textbf{\PSMr} is the SOTA optimization-based method that requires a complete surface shape representation. We use its open-source implementation in ShapeWorks\cite{cates2017shapeworks}. It is included to provide a reference point for the SSM metrics.
    \item \textbf{\DAEr} is the autoencoder formulated by \cite{achlioptas2018learning} with the same DGCNN \cite{wang2019dgcnn} encoder as \PS.
    \item \textbf{\ISRr} is the method proposed for learning intrinsic structural representation (ISR) points via  PointNet++ \cite{qi2017pointnet++} encoder and an MLP point integration module. 
    \item \textbf{\DPCr} is the pairwise deep point correspondence (DPC) model that reorders a source point cloud to match a target point cloud via latent similarity. In inference, we set the target as the point cloud with the minimum CD to all others to acquire population-level correspondence. 
    \item \textbf{\SCr} is the Self-supervised and Coherent 3D Keypoints (SC3K) estimation method proposed in concurrence with Point2SSM. This method detects sparse keypoints from point clouds that are noisy, down-sampled, and arbitrarily rotated. 
    \item \textbf{\SCPr} denotes the \SC model with our \PSP loss (\eqref{eq:consist_loss}). The original \SC loss was proposed for sparse output points ($\M \approx 10$) and fails given denser output ($\M > 50$) \cite{zohaib2023sc3k}. Thus, we include \SCP to compare the \SC and \PSP architectures alone. 
\end{itemize}

\edit{For completeness, we additionally report a supplementary comparison with a deformation-based statistical shape model (FlowSSM \cite{ludke2022landmark}) \ref{app:flowssm}. This comparison requires additional preprocessing and is therefore presented separately from the main point-cloud–based results.}


\subsubsection{Results}

\Figref{fig:benchmark} summarizes the results across metrics (\secref{sec:metrics}), datasets (\secref{sec:datasets}), and models (\secref{sec:models}).
\PSP provides the best overall accuracy, outperforming the other point cloud deep learning-based methods in terms of surface sampling, consistency, and correspondence.
\revise{\ref{app:single_pvalues} provides additional statistical analyses of these results, demonstrating that the error reductions achieved by \PSP are statistically significant in most cases.}  
Additionally, \PSP matches the traditional \PSM model performance on SSM metrics, achieving similar or better compactness and better generalization and specificity while significantly relaxing the input shape requirement.
 
\begin{figure}[!t]
    \centering
    \includegraphics[width=.48\textwidth]{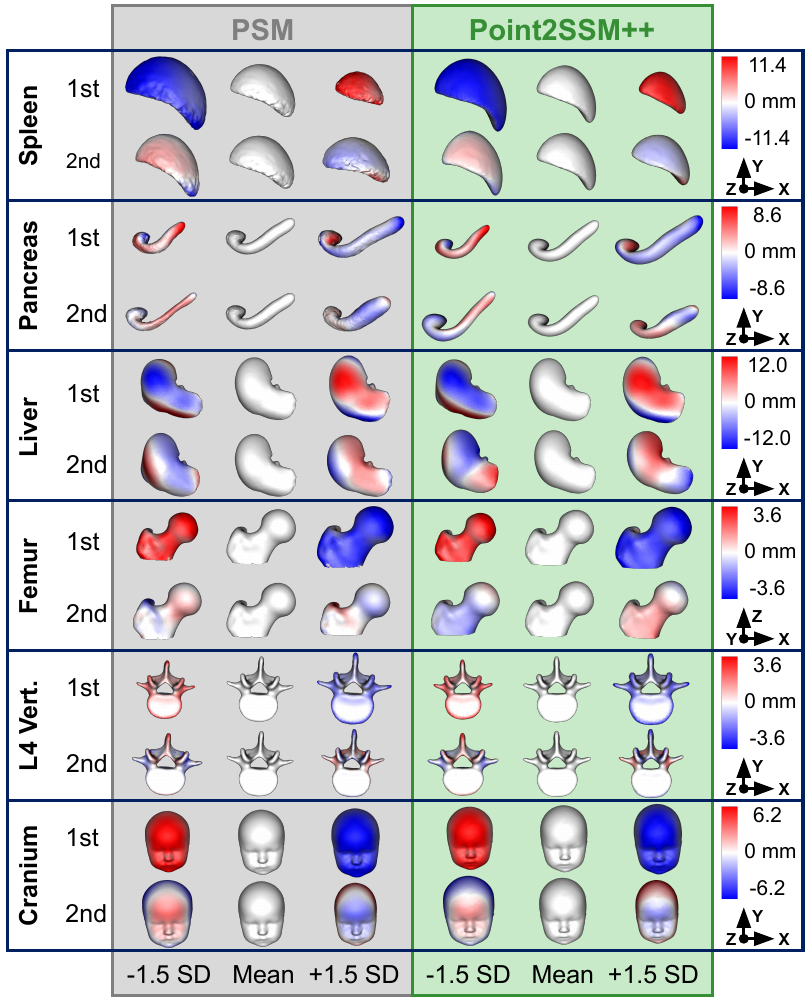}%
    \caption{Primary and secondary modes of population variation captured by the \PSM (gray) and \PSP (green) methods across each dataset. The mean and shapes at $\pm 1.5$ standard deviation are shown with color denoting the distance to the mean. Blue indicates the surface is outside the mean, and red indicates the surface is inside the mean. Separate color scales and orientation markers are provided for each dataset. }
    \label{fig:benchmark_modes}
\end{figure}

\Figref{fig:benchmark_modes} shows the primary and secondary modes of variation captured by the \PSM and \PSP methods for each of the datasets. A similar visualization is provided for all comparison models in \ref{app:modes}. The mean and shapes at $\pm 1.5$ standard deviation are displayed as meshes, generated via mesh warping using the output correspondence points as control points. \PSP provides similar, if not more, smooth and interpretable mean shapes and modes of variation as \PSM. The mean shapes resulting from \PSM on datasets with a large amount of linearity (\ie spleen and pancreas) have some artifacts indicating some miscorreposondence impacting the mesh warping. The \PSP mean shapes do not have this issue. In most cases, the primary modes found by the two methods are similar, with size often captured as the first mode. This result suggests that \PSP is correctly capturing the population-level statistics of each shape cohort.

\begin{figure}[!th]
    \centering
    \includegraphics[width=.48\textwidth]{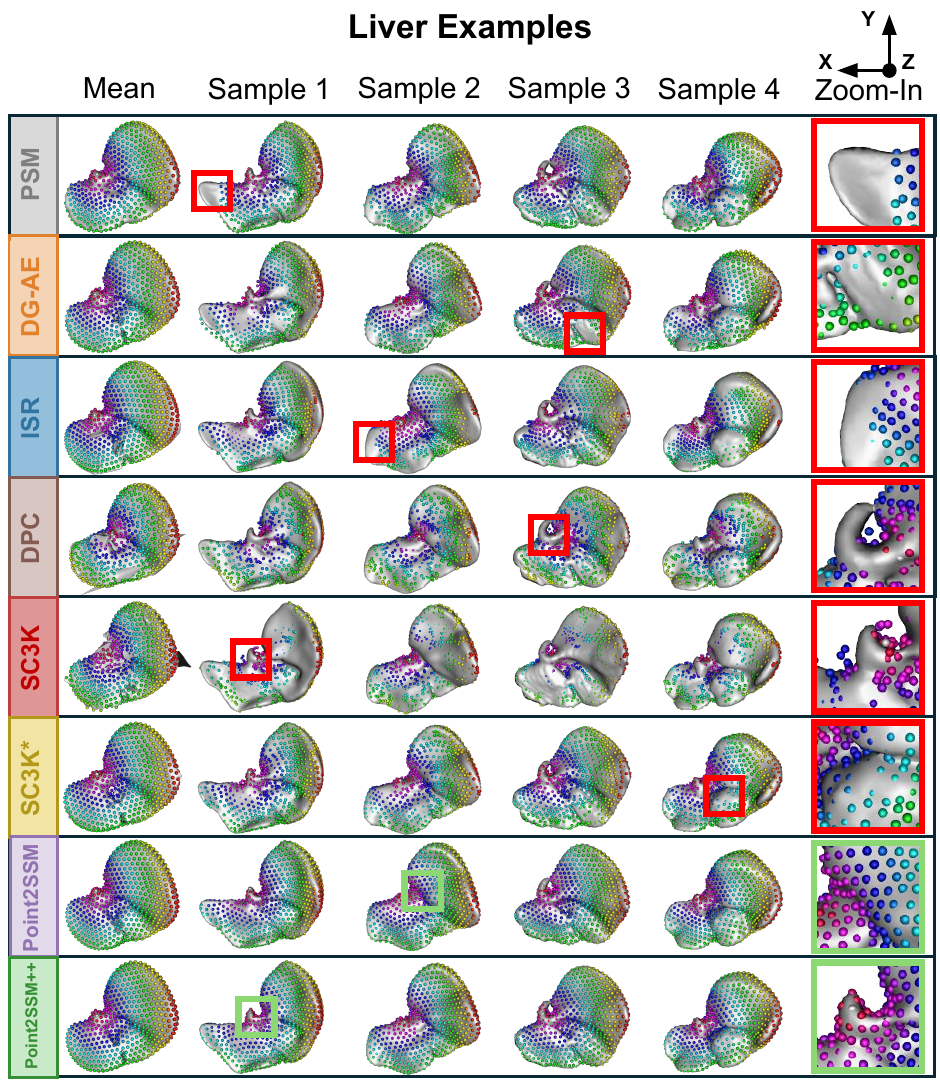}%
    \caption{The mean correspondence points and four example predictions from the test set are provided from each approach on the liver dataset. The mean mesh is constructed via mesh warping. Sample correspondence point predictions are overlaid over ground truth meshes. Color denotes correspondence. An illustrative zoom-in box is provided for each model.}
    \label{fig:liver}
\end{figure}

\Figref{fig:liver} provides qualitative results on the liver dataset. The mean correspondence points from each model are displayed, as well as four example predictions from the test set. The mean correspondence points should represent a plausible liver. Mean meshes are reconstructed via correspondence-based mesh warping, thus mesh artifacts suggest miscorrespondence. \PSP provides a plausible mean shape with a smooth mesh, whereas the mean meshes resulting from other methods (such as \DPC and \SC) have artifacts. 
All methods establish reasonable correspondence, as can be seen by the color of the points in \figref{fig:liver}. However, not every method provides good surface sampling, as shown in the red zoomed-in boxes which highlight predicted points that do not lie on the true surface. Even in the case of the traditional \PSM method, where points are directly constrained to the surface, the points do not cover the entire shape due to the high variability of the liver shape.
\PSP provides the best correspondence with uniformly spread points, constrained to the surface in semantically consistent locations across the samples.

\revise{A training time comparison is provided in \ref{app:runtime}. \PSP requires a similar training time as the other point cloud deep learning models and does not substantially increase runtime over the base \PS model. On the largest dataset (cranium), \PSP and the other deep models required roughly an hour and a half to train, whereas the \PSM optimization took more than a day. Although inference with \PSP requires two samples, the test runtime remains very fast, requiring under a second per shape.}

\subsubsection{Downstream Task Analysis}

\begin{figure}[!h]
    \centering
    \includegraphics[width=.48\textwidth]{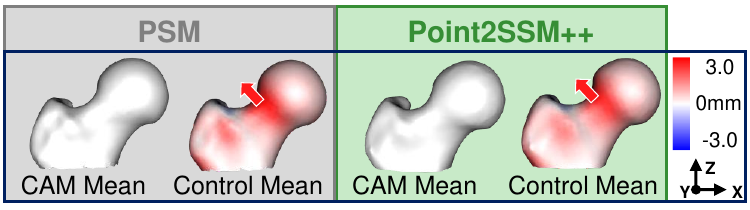}
    \caption{Femur group differences from \PSM and \PSP. CAM pathology and control group mean shapes are displayed. Color denotes the distance from the control mean to the CAM mean. The pathology is correctly captured as a widening of the femoral neck.}
    \label{fig:femur}
\end{figure}

The femur and cranium datasets have subgroups, enabling downstream clinical task analysis. 
The femur dataset contains 47 healthy/control subjects and 9 with CAM pathology, characterized by an abnormal bone growth lesion on the femoral neck that limits motion and is associated with hip osteoarthritis \cite{atkins2017quantitative}.
The \PSP model captures this difference similarly to the traditional \PSM method, as seen by the colormap distance between the means in \figref{fig:femur}. This experiment serves as an example of how \PSP can be used to characterize pathology.

\begin{figure}[h!]
    \centering
    \includegraphics[width=.48\textwidth]{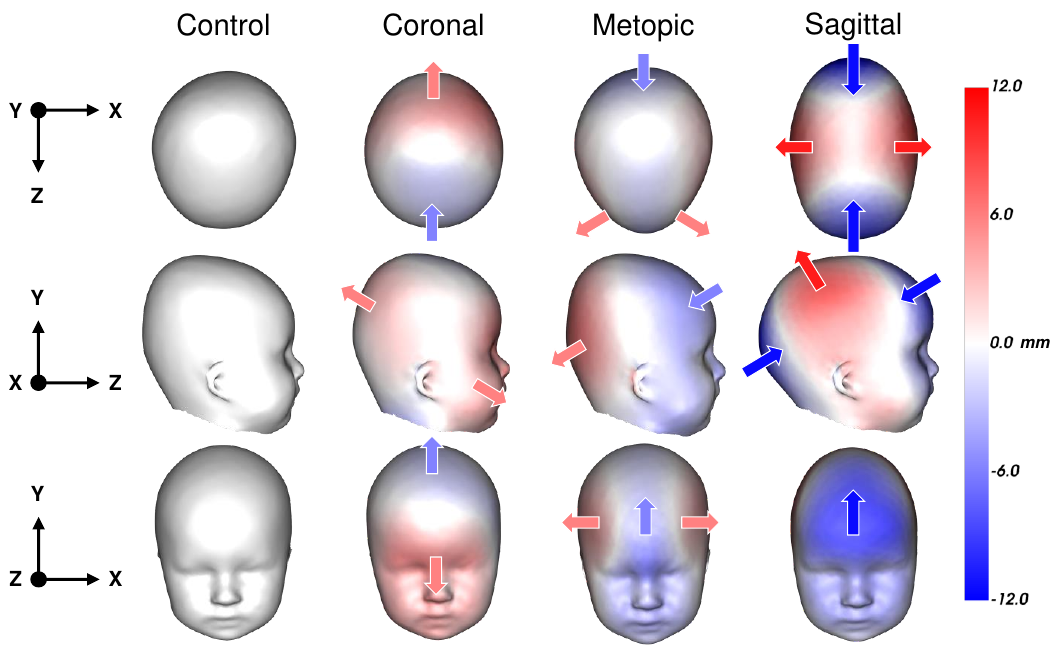}
    \caption{Cranio group differences from \PSP. The mean shape of each group resulting from \PSP predicted SSM is shown from multiple views. Color denotes the distance of the shapes to the control mean shape. Arrow annotations are provided to show the general trends. }
    \label{fig:cranio}
\end{figure}

The cranium dataset contains control samples and three categories of craniosynostosis, a disorder in which one or more of the sutures between the bones of infants' cranial sutures fuse prematurely. In \figref{fig:cranio}, we utilize the \PSP output to compare the mean shape of each of these subgroups to the control group. Each of the subgroups is correctly characterized, with coronal synostosis manifesting as a flattening of the forehead, metopic as a triangulation of the forehead, and sagittal as a narrowing of the skull. 

SSM on the cranium dataset could be useful in classifying new shapes as normal cases or as one of the three types of craniosynostosis. We trained a classifier on SSM predictions from each construction technique to explore their effectiveness for classification. In particular, first, correspondences were obtained on training and testing samples using each model. Then, we trained a support vector classification model with RBF kernels on the training correspondences and evaluated the classification accuracy on test correspondences. We utilized five fold cross-validation to repeat this experiment five times with different train/test splits to ensure fair analysis. Note the classes are balanced in this dataset and in all splits.

\begin{table}[!h]
\caption{Craniosynostosis classification accuracy: Mean values of ratio of correctly classified instances and class F1 scores are with standard deviation from five fold cross-validation.\label{tab:cranio}}
\centering
\resizebox{.48\textwidth}{!}{%
\begin{tabular}{cc|cccc|}
\cline{3-6}
 &  & \multicolumn{4}{c|}{Class F1 Score $\uparrow$} \\ \hline
\multicolumn{1}{|c|}{Model} & Accuracy $\uparrow$ & \multicolumn{1}{c|}{Control} & \multicolumn{1}{c|}{Coronal} & \multicolumn{1}{c|}{Metopic} & Sagittal \\ \hline\hline

\multicolumn{1}{|c|}{\PSM} & 0.965$\pm$0.015 & \multicolumn{1}{c|}{0.943$\pm$0.032}  & \multicolumn{1}{c|}{0.963$\pm$0.032} & \multicolumn{1}{c|}{0.985$\pm$0.012} & {0.969$\pm$0.020} \\ \hline

\multicolumn{1}{|c|}{\DAE} & 0.977$\pm$0.005 & \multicolumn{1}{c|}{0.975$\pm$0.017}  & \multicolumn{1}{c|}{0.995$\pm$0.010} & \multicolumn{1}{c|}{0.980$\pm$0.019} & {0.959$\pm$0.013} \\ \hline

\multicolumn{1}{|c|}{\ISR} & 0.930$\pm$0.026 & \multicolumn{1}{c|}{0.902$\pm$0.039}  & \multicolumn{1}{c|}{0.950$\pm$0.035} & \multicolumn{1}{c|}{0.934$\pm$0.041} & {0.933$\pm$0.037} \\ \hline

\multicolumn{1}{|c|}{\DPC} & 0.953$\pm$0.028 & \multicolumn{1}{c|}{0.941$\pm$0.025}  & \multicolumn{1}{c|}{0.964$\pm$0.036} & \multicolumn{1}{c|}{0.937$\pm$0.028} & {0.969$\pm$0.039} \\ \hline

\multicolumn{1}{|c|}{\SC} & 0.970$\pm$0.023 & \multicolumn{1}{c|}{0.968$\pm$0.052}  & \multicolumn{1}{c|}{0.980$\pm$0.019} & \multicolumn{1}{c|}{0.960$\pm$0.019} & {0.970$\pm$0.019} \\ \hline

\multicolumn{1}{|c|}{\SCP} & 0.970$\pm$0.010 & \multicolumn{1}{c|}{0.960$\pm$0.021}  & \multicolumn{1}{c|}{0.980$\pm$0.010} & \multicolumn{1}{c|}{0.970$\pm$0.019} & {0.969$\pm$0.020} \\ \hline

\multicolumn{1}{|c|}{\PS} & \textbf{0.982$\pm$0.006} & \multicolumn{1}{c|}{\textbf{0.985$\pm$0.021}}  & \multicolumn{1}{c|}{\textbf{0.985$\pm$0.012}} & \multicolumn{1}{c|}{0.985$\pm$0.012} & \textbf{{0.974$\pm$0.017}} \\ \hline

\multicolumn{1}{|c|}{\PSP} & \textbf{0.982$\pm$0.006} & \multicolumn{1}{c|}{\textbf{0.985$\pm$0.012}}  & \multicolumn{1}{c|}{\textbf{0.985$\pm$0.012}} & \multicolumn{1}{c|}{\textbf{0.990$\pm$0.012}} & {0.969$\pm$0.020} \\ \hline
\end{tabular}%
}
\end{table}

The results are summarized in \tabref{tab:cranio}. All SSMs correctly classified at least $93\%$  of the cases, but \PS and \PSP provided the best accuracy at $98.2\%$ correct. These results further indicate \PSP provides SSM that correctly captures the morphology indicative of the subtypes of craniosynostosis.

\newpage
\subsection{Multi-Anatomy Experiments}
\label{sec:multianatomy}

We utilize three models in multi-class experiments:
\begin{itemize}
    \item \textbf{Single-Class Point2SSM++} denotes the model trained and tested on only a single anatomy, as was done in \secref{sec:benchmark}.
    \item \textbf{Multiclass Point2SSM++} denotes the model trained on multiple anatomies.
    \item \textbf{Multiclass Point2SSM++, Classifier} denotes the multi-anatomy \PSP model with an auxiliary classifier as explained in \secref{sec:multianatomy_methods}.
\end{itemize}
The following sections detail multi-anatomy experiment results on two different datasets.

\subsubsection{Multi-Organ Experiment}
The multi-organ dataset is comprised of three anatomy types: the spleen, pancreas, and liver datasets described in \secref{sec:datasets}. 
We randomly select a subset of 32 training shapes for each organ, resulting in a total training set of 96. The single anatomy \PSP models are individually trained on each subset of 32 shapes. This experiment provides an example of multi-anatomy training with large differences between anatomy types. 

\begin{figure}[!ht]
    \centering
    \includegraphics[width=.48\textwidth]{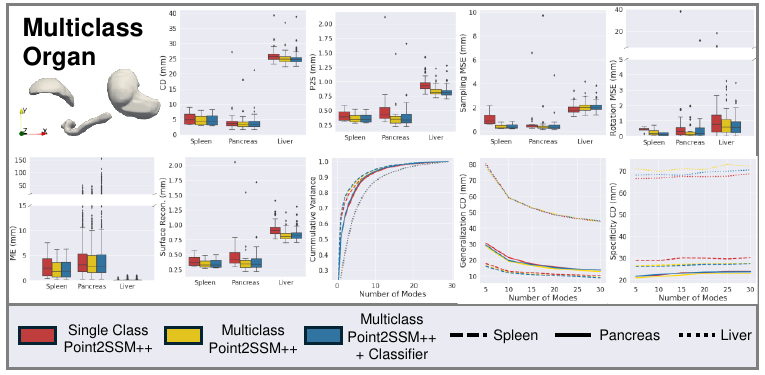}%
    \caption{Multi-organ results. Metrics are reported for the three models across the three organs \revise{(see \ref{app:multi_pvalues} for analysis of statistical significance)}. Line types in the PCA plots differentiate organs.}
    \label{fig:multi_organ}
\end{figure} 

The results are compiled in Figure \ref{fig:multi_organ}. The multi-anatomy \PSP models achieve the best performance \revise{with statistically significant improvements in error over the single-class models (see \ref{app:multi_pvalues} for statistical analysis)}. These results demonstrate that by increasing the training set size via multi-anatomy training, the model is made more robust. This experiment suggests that an anatomy-agnostic multi-anatomy \PSP could be trained on a wide variety of shapes to provide a foundational SSM model.

\subsubsection{Vertebrae Experiment}
\begin{figure}[h!]
    \centering
    \includegraphics[width=.48\textwidth]{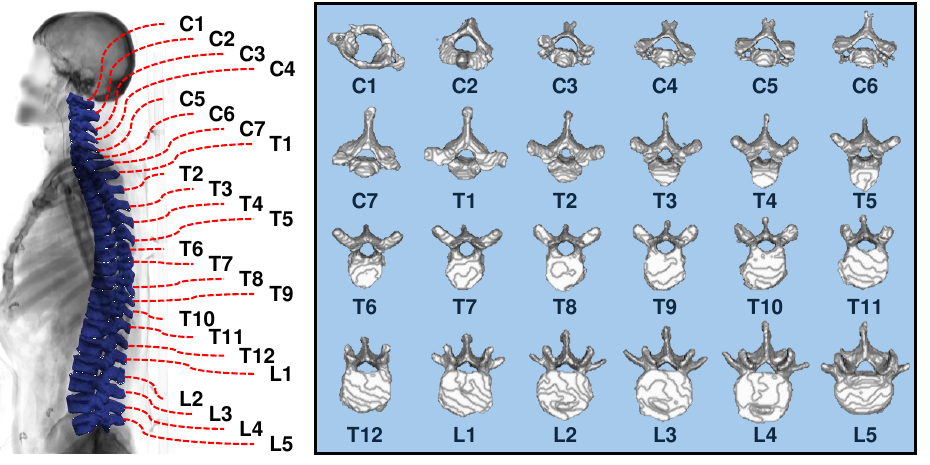}
    \caption{Example of the 24 vertebrae bones from side and top view.}
    \label{fig:vert}
\end{figure}

The vertebrae dataset includes 24 vertebrae bones from the TotalSegmentator dataset \cite{wasserthal2023totalsegmentator}. This includes cervical (C1-C7), thoracic (T1-T12), and lumbar spine (L1-L5). We randomly select 30 training, 30 validation, and 100 test shapes for each bone to create this dataset.
The bones have shared shape properties but vary significantly between classes, as can be seen in \figref{fig:vert}.
Additionally, the bone segmentations in this dataset are binary mask volumes with varying resolution and often large $z$-spacing, many of which have artifacts like holes and islands. Thus, acquiring complete surface meshes is difficult. We utilize marching cubes to extract point clouds from the binary volumes for \PSP input. This experiment reinforces the versatility of \PSP, as it can be used given imperfect input.

\begin{figure}[h!]
    \centering
    \includegraphics[width=.48\textwidth]{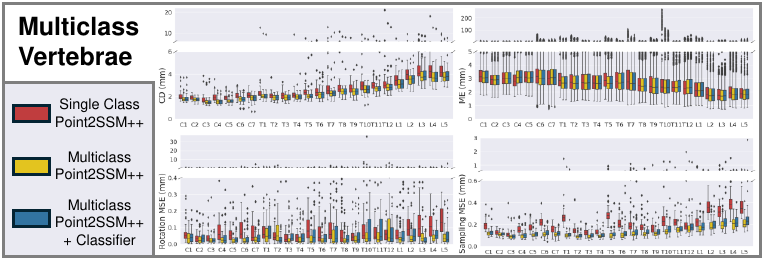}
    \caption{Multi-anatomy vertebrae results. Metrics are reported for the three models across the 24 vertebrae \revise{(see \ref{app:multi_pvalues} for analysis of statistical significance)}.}
    \label{fig:multi_vert}
\end{figure}

Evaluation metrics for each of the 24 vertebrae are provided in \figref{fig:multi_vert}. Note the P2S and Warp P2P metrics are excluded here due to the lack of ground truth surface meshes. The multi-anatomy models perform similarly or better than the single anatomy model across all vertebrae. 
SSM metric plots would be difficult to interpret with all bones, and thus summary values are reported in \tabref{tab:vert}. Here, compactness is quantified as the number of modes required to capture $95\%$ of the variation, and generalization and specificity are reported with that number of modes. The single anatomy and multi-anatomy models provide similar PCA statistics. 

\begin{table}[!h]
\caption{Multiclass Vertebrae SSM Metrics\label{tab:vert}}
\centering
\resizebox{.48\textwidth}{!}{%
\begin{tabular}{|l|ccc|ccc|ccc|}
\hline
\multicolumn{1}{|l|}{\multirow{2}{*}{Vert.}} & \multicolumn{3}{c|}{ Compactness $\downarrow$} & \multicolumn{3}{c|}{Generalization CD (mm) $\downarrow$} & \multicolumn{3}{c|}{Specificity CD (mm) $\downarrow$} \\ \cline{2-10} 
\multicolumn{1}{|l|}{} & Single & Multi & \multicolumn{1}{l|}{\makecell{Multi + \\ Classifier}} & Single & Multi & \multicolumn{1}{l|}{\makecell{Multi + \\ Classifier}} & Single & Multi & \multicolumn{1}{l|}{\makecell{Multi + \\ Classifier}} \\ \hline

C1 & 18 & 17 & 17 & 3.12 & 3.09 & 3.1 & 3.38 & 3.34 & 3.34 \\
C2 & 14 & 14 & 14 & 2.76 & 2.75 & 2.75 & 3.04 & 3.04 & 3.09 \\
C3 & 18 & 17 & 18 & 2.68 & 2.64 & 2.64 & 3.13 & 3.15 & 3.15 \\
C4 & 18 & 17 & 17 & 2.9 & 2.78 & 2.75 & 3.5 & 3.44 & 3.47 \\
C5 & 15 & 14 & 14 & 2.92 & 2.87 & 2.86 & 3.83 & 3.96 & 3.87 \\
C6 & 13 & 13 & 13 & 3.32 & 3.31 & 3.28 & 3.52 & 3.59 & 3.54 \\
C7 & 8 &  8 &  8 &  3.7 & 3.73 & 3.7 & 3.4 & 3.59 & 3.5 \\
T1 & 15 & 13 & 13 & 3.7 & 3.6 & 3.64 & 3.67 & 3.63 & 3.6 \\
T2 & 12 & 11 & 12 & 2.97 & 3.03 & 3.01 & 3.62 & 3.58 & 3.66 \\
T3 & 13 & 13 & 13 & 3.0 & 3.03 & 2.97 & 3.23 & 3.32 & 3.31 \\
T4 & 10 & 9 &  10 & 3.37 & 3.5 & 3.43 & 3.55 & 3.55 & 3.56 \\
T5 & 13 & 12 & 13 & 3.63 & 3.54 & 3.53 & 3.8 & 3.81 & 3.84 \\
T6 & 13 & 11 & 11 & 4.12 & 4.14 & 4.2 & 3.82 & 3.77 & 3.8 \\
T7 & 10 & 10 & 11 & 6.13 & 6.38 & 6.33 & 3.94 & 4.04 & 4.07 \\
T8 & 9 &  8 &  9 &  4.6 & 4.75 & 4.62 & 3.91 & 3.92 & 3.98 \\
T9 & 12 & 11 & 11 & 4.33 & 4.18 & 4.22 & 4.66 & 4.6 & 4.6 \\
T10 & 14 & 13 & 14 & 5.17 & 4.93 & 5.0 & 4.53 & 4.52 & 4.5 \\
T11 & 17 & 15 & 16 & 4.54 & 4.48 & 4.44 & 4.74 & 4.82 & 4.74 \\
T12 & 18 & 16 & 16 & 5.4 & 5.39 & 5.47 & 4.67 & 4.74 & 4.7 \\
L1 & 17 & 16 & 16 & 5.61 & 5.48 & 5.44 & 5.17 & 5.16 & 5.23 \\
L2 & 16 & 15 & 16 & 5.68 & 5.47 & 5.5 & 5.59 & 5.44 & 5.6 \\
L3 & 11 & 10 & 10 & 6.91 & 6.96 & 7.01 & 7.01 & 7.09 & 6.97 \\
L4 & 13 & 13 & 13 & 7.73 & 7.64 & 7.69 & 6.72 & 6.72 & 6.69 \\
L5 & 15 & 15 & 14 & 7.18 & 7.19 & 7.21 & 6.59 & 6.66 & 6.5 \\ \hline
Avg. & 13.83 & 12.96 & 13.29 & 4.39 & 4.37 & 4.37 & 4.29 & 4.31 & 4.30 \\ \hline
\end{tabular}%
}
\end{table} 

\begin{figure}[ht!]
    \centering
    \includegraphics[width=.48\textwidth]{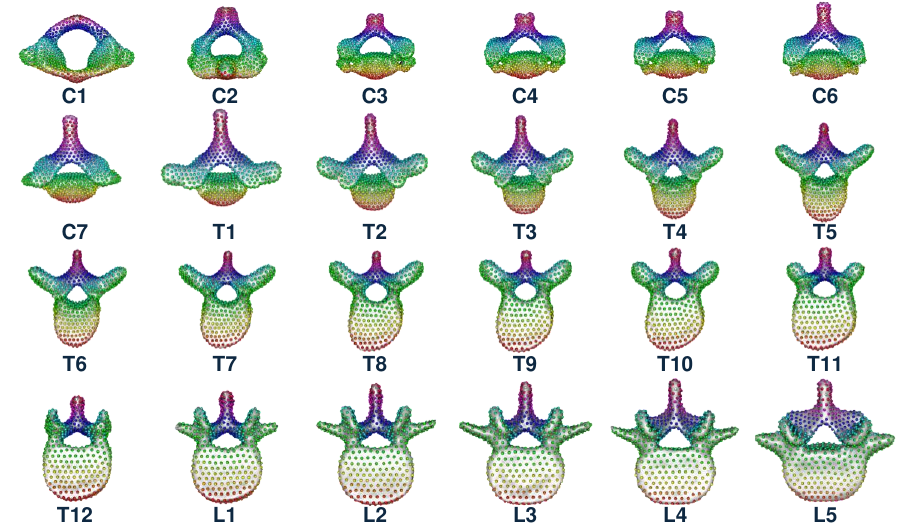}
    \caption{The mean predictions from the multi-anatomy \PSP, Classifier model for each vertebrae. Color denotes correspondence.}
    \label{fig:vert_means}
\end{figure}

\Figref{fig:vert_means} shows that the multi-anatomy \PSP model with the auxiliary classifier model provides correspondence across different vertebrae.
This visualization highlights the utility of the approach, as it can be used to quantify and characterize shape differences not only within each anatomy but also across anatomy types. 

\begin{table*}[h!]
\caption{Multiclass Vertebrae Classification Accuracy \label{tab:vert_class}}
\centering
\resizebox{\textwidth}{!}{%
\begin{tabular}{|l|c|cccccccccccccccccccccccc|} \hline
\multirow{2}{*}{Model} & \multirow{2}{*}{\makecell{Overall\\Acurracy}} & \multicolumn{24}{c|}{Individual Vertebrae F1 Scores} \\
 &  & C1 & C2 & C3 & C4 & C5 & C6 & C7 & T1 & T2 & T3 & T4 & T5 & T6 & T7 & T8 & T9 & T10 & T11 & T12 & L1 & L2 & L3 & L4 & L5  \\ \hline
Multi & 0.6533 & \textbf{1.00} & \textbf{1.00} & 0.60 & 0.37 & 0.24 & 0.73 & \textbf{0.86} & 0.91 & 0.78 & 0.54 & 0.55 & 0.43 & 0.45 & 0.38 & \textbf{0.41} & 0.55 & 0.64 & \textbf{0.84} &
\textbf{0.89} & 0.59 & 0.45 & 0.63 & 0.78 &	0.93
\\
Multi+Class. & \textbf{0.6825} & \textbf{1.00} & \textbf{1.00} & \textbf{0.62 }& \textbf{0.42} &\textbf{ 0.28} & \textbf{0.80} & \textbf{0.86} & \textbf{0.92} & \textbf{0.79} & \textbf{0.61} & \textbf{0.63} & \textbf{0.50} & \textbf{0.48} & \textbf{0.45} & 0.36 & \textbf{0.60} & \textbf{0.68} & 0.82 & 0.88 & \textbf{0.61} &\textbf{0.46} & \textbf{0.67} &	\textbf{0.81} & \textbf{0.94}
\\ \hline
\end{tabular}%
}
\end{table*}

The motivation behind including an auxiliary classifier is to promote learning anatomy-specific features. Thus, distinguishing between anatomy types in the output correspondence points should be easier. To evaluate this, we trained a classifier on the predictions made by the multi-anatomy \PSP with and without the auxiliary classifier. In particular, we fit a 10-nearest neighbors classifier to the predicted training correspondence points and evaluate the accuracy of the predicted test correspondence points. The results are provided in \tabref{tab:vert_class}. The multi-anatomy model alone achieved an accuracy of $65.33\%$ whereas the multi-anatomy model with the auxiliary classifier achieved an accuracy of $68.25\%$. This difference suggests that while the models achieve similar performance on the metrics reported in \figref{fig:multi_vert} and \tabref{tab:vert}, the model with the auxiliary classifier does better in capturing anatomy-specific features. 

\subsection{Spatiotemporal (4D) Experiments}
\label{sec:4d}
Next, we evaluate the ability to adapt \PSP to the case spatiotemporal/4D SSM where we have sequences of shape from a subject over time for dynamic or longitudinal clinical studies. 
The models evaluated on the task of spatiotemporal anatomical SSM are outlined as follows:
\begin{itemize}
    \item \textbf{Spatiotemporal PSM \cite{adams2022polynomial4d}} denotes the 4D extension of the optimization-based PSM method. This approach injects the underlying time dependency into correspondence optimization via a regularized principal component polynomial regression across time. Like \PSM, it requires a complete surface shape representation but is included to provide context for the SSM metrics.
    \item \textbf{3D \PSP} denotes the proposed \PSP approach applied cross-sectionally to 4D data, ignoring the time dependency. This baseline is included to highlight the importance of modeling the time dependency. 
    \item \textbf{4D PSTNet2 \PSP} denotes the \PSP method where the 3D DGCNN encoder is replaced by the hierarchical Point Spatio-Temporal net (PSTnet2) \cite{fan2021pstnet2}, as explained in \secref{sec:4d_methods}.
\end{itemize}

\subsubsection{4D Ellipsoid Experiment}

\begin{figure}[!h]
    \centering
    \includegraphics[width=.48\textwidth]{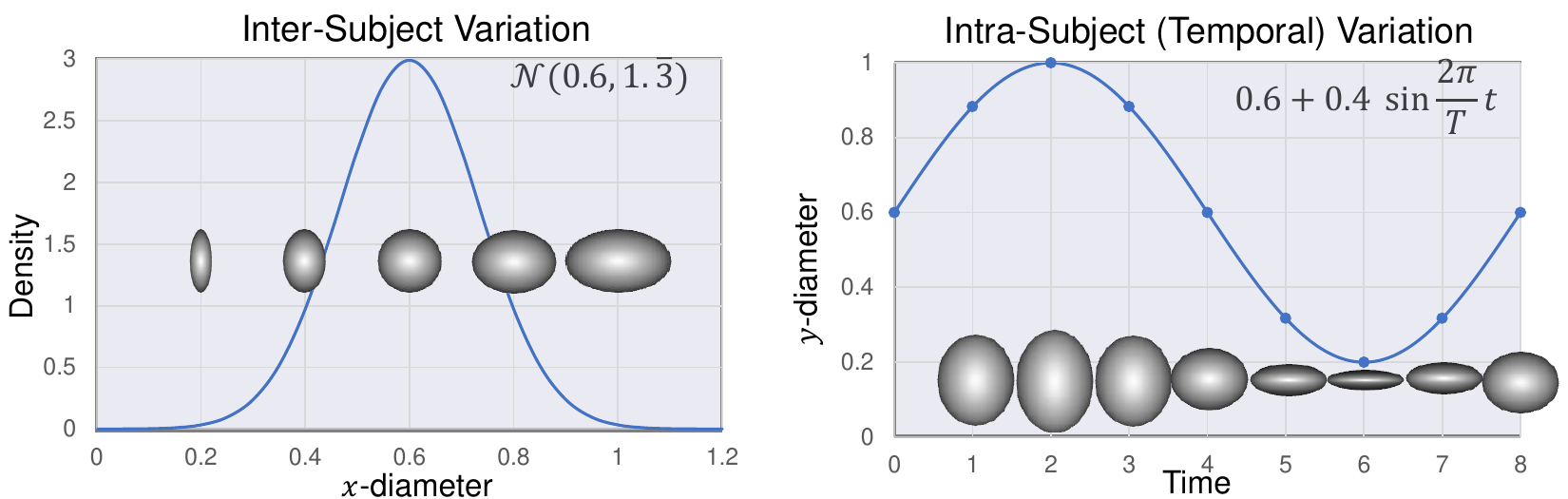}
    \caption{The functions defining the diameters of the synthetic 4D ellipsoid dataset are shown with example shapes. The $x$-diameter provides inter-subject variation, and the $y$-diameter provides intra-subject variation across time. }
    \label{fig:ellipsoid}
\end{figure}

We utilize a synthetic 4D ellipsoid dataset with known shape dynamics as proof of concept. These ellipsoids are parameterized by a subject-specific $x$-diameter (randomly selected from a Gaussian distribution), a time-dependent $y$-diameter, and a fixed population consistent $z$-diameter value, as summarized in \figref{fig:ellipsoid}.

\begin{figure}[!h]
    \centering
    \includegraphics[width=.48\textwidth]{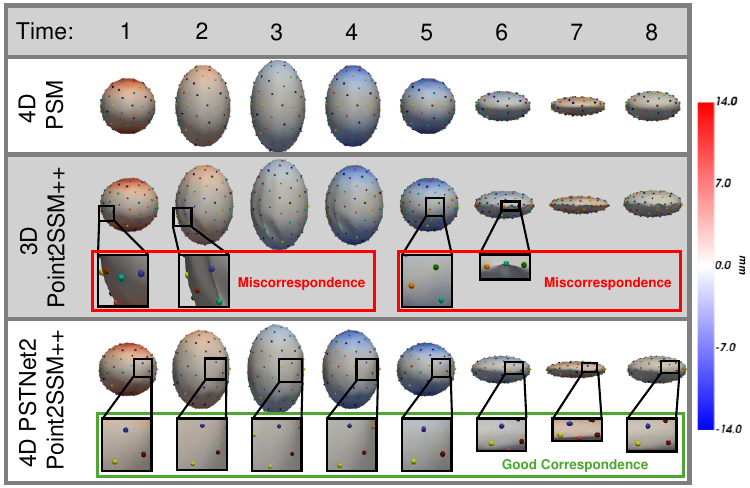}%
    \caption{Mean correspondence points resulting from each model are shown over reconstructed meshes. Point color denotes correspondence. Mesh color denotes the distance to the next time point mean.}
    \label{fig:4d_ellipsoids}
\end{figure}

\Figref{fig:4d_ellipsoids} shows the predicted mean correspondence points across time resulting from each of the models. 
All three methods correctly find the significant modes of variation, meaning two significant modes across subjects and time ($x$ and $y$ diameter), one significant mode across subjects within each time point ($x$-diameter), and one significant mode across time within each subject's sequence ($y$-diameter). 
However, as can be seen in \figref{fig:4d_ellipsoids}, the 4D PSTNet2 \PSP model provides better correspondence than the 3D \PSP method applied cross-sectionally. This proof of concept experiment demonstrates the benefit of using a 4D encoder in \PSP with spatiotemporal datasets.

\subsubsection{4D Left Atrium Experiment}
Next, we analyze the accuracy of the 4D PSTNet2 \PSP method compared to the traditional approach, 4D \PSM, on a real spatiotemporal anatomical dataset: the left atrium of the heart. 
This dataset includes 3D late gadolinium enhancement (LGE) and stacked cine cardiovascular magnetic resonance (CMR) scans from 28 patients with atrial fibrillation, gathered prior to radiofrequency ablation procedures. Each cine scan consisted of 25 time points covering a single cardiac cycle (between R-wave peaks), with temporal normalization during acquisition to encompass one heartbeat per patient. 
The 3D LGE images were segmented manually by a cardiac imaging specialist and aligned with the nearest cine time point based on CMR trigger time. This segmentation was then propagated to all time points using pairwise deformable registrations, resulting in a complete 3D segmentation for each time point \cite{morris2020image}.  

\begin{figure}[!ht]
    \centering
    \includegraphics[width=.48\textwidth]{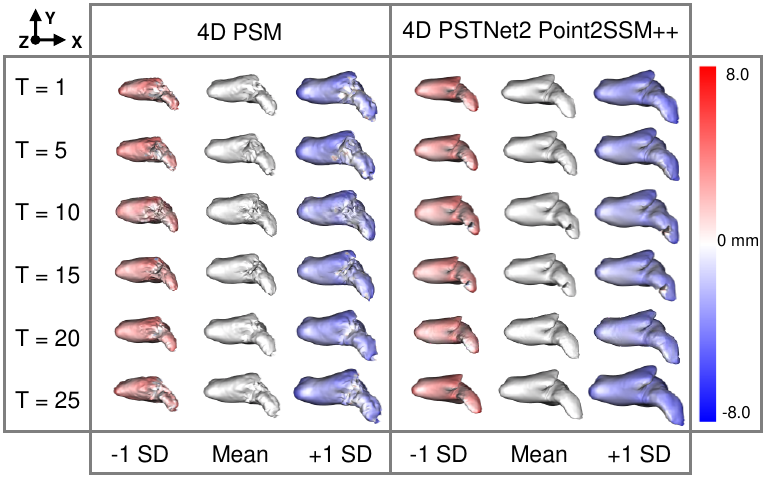}%
    \caption{The primary mode of variation resulting from the 4D \PSM method and 4D PSTNet2 \PSP on the left atrium at various time points is shown from the top view. Color denotes the distance to the mean shape.}
    \label{fig:4D_LA}
\end{figure}

\Figref{fig:4D_LA} shows the mean shape and primary mode of variation captured across time by 4D \PSM and 4D PSTNet2 \PSP. The mean shape and variation are similar; however, the 4D PSTNet2 \PSP mesh constructions have fewer artifacts in the pulmonary veins and left atrium appendage. This improvement indicates that the 4D PSTNet2 \PSP provides better correspondence. 

\begin{figure}[!ht]
    \centering
    \includegraphics[width=.48\textwidth]{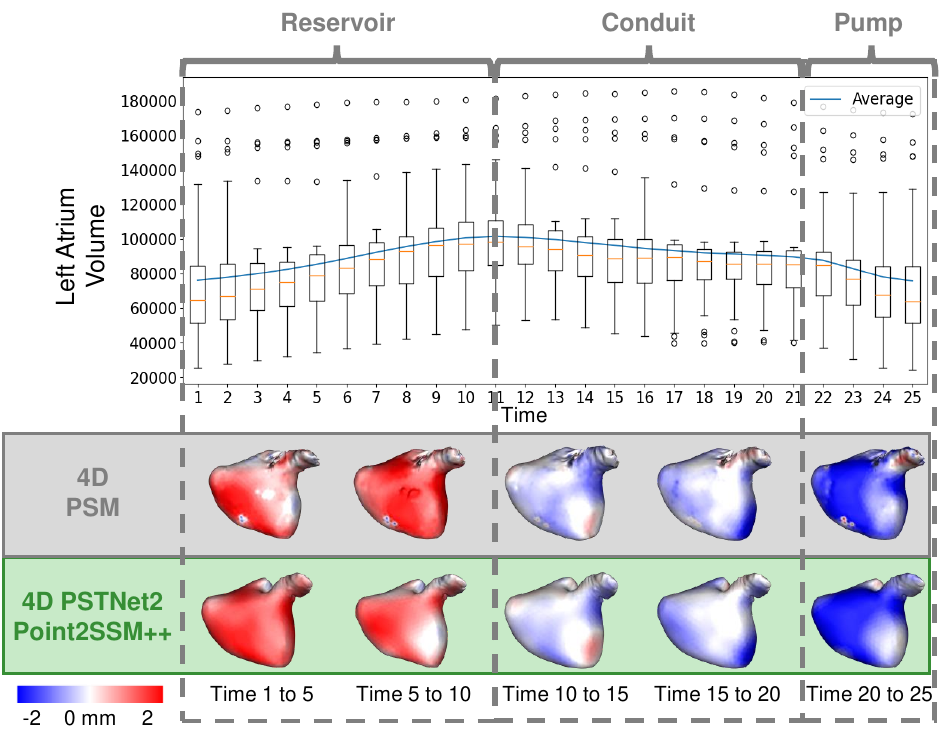}%
    \caption{The three left atrium function intervals are shown: reservoir, conduit, and pump. The top plot displays box plots of the left atrium volume across time. At the bottom the mean shapes from 4D \PSM and 4D PSTNet2 \PSP are shown from the anterior view at a subset of time points. Heat maps show the change in shape to the next displayed time point mean shape (red denotes expansions and blue contraction).}
    \label{fig:LA_volume}
\end{figure}

The dynamics of the left atrium during a heartbeat are characterized by three steps: reservoir or filling (where the volume increases), conduit or passive emptying (where volume decreases slowly in a decelerating manner), and pump or active emptying (where volume decreases quickly). 
\Figref{fig:LA_volume} shows that both methods correctly characterize this temporal dependency, as the mean shape dynamics correctly match the mean volume over time.
However, as can be seen by the mean colormaps \figref{fig:LA_volume}, 4D PSTNet2 \PSP provides a smoother characterization of the dynamics, indicating better correspondence. These results demonstrate that \PSP can easily be adapted to accurately capture both spatial and temporal variation in anatomical cohorts, enabling longitudinal and dynamic clinical studies. 
\section{Conclusion}
\label{sec:conclusion}

In this paper, we introduced \PSP, a principled, self-supervised approach for constructing correspondence-based anatomical SSM directly from point clouds. Through extensive evaluation, we demonstrated \PSP outperforms existing deep learning-based methods while eliminating the need for preprocessing steps such as 
\edit{surface reconstruction, segmentation, or point cloud alignment.}
Moreover, our results indicate that \PSP matches or surpasses the performance of the optimization-based \PSM method, while alleviating the associated burdens, including stringent input shape requirements, susceptibility to bias/assumptions, and prolonged inference processes. Additionally, our experiments involving multiclass and 4D spatiotemporal analyses showcase the versatility of \PSP across various clinical tasks.

\edit{Recent deformation-based SSMs, including flow- and diffeomorphism-based models, offer a powerful alternative paradigm when a template surface and dense parameterization are available. These methods enable smooth transformations and flexible landmark propagation within a canonical shape space. In contrast, \PSP is designed for settings in which such preprocessing is impractical or unavailable, operating directly on unordered point clouds with minimal assumptions. We therefore view correspondence-learning approaches such as \PSP and deformation-based SSMs as complementary, each addressing distinct data regimes and application requirements. Supplementary experiments in Appendix \ref{app:flowssm} comparing \PSP with flow-based SSMs further support this perspective.}
 
Potential avenues for future research include exploring probabilistic extensions of \PSP to provide estimations of uncertainty or model confidence, ensuring the reliability of predictions in clinical settings. As \PSr has previously demonstrated robustness to imperfect input such as noisy, sparse, and partial point clouds, future investigations could also focus on directly enhancing this inherited robustness in \PSP. Further experiments could be conducted with challenging clinical shape representations, such as sparse binary segmentation masks with large slice distances.

Overall, \PSP significantly streamlines SSM construction and inference by removing all manual and automatic preprocessing steps and reducing computational overhead and time requirements. This improvement paves the way for SSM to become a widely accessible onsite clinical tool. Furthermore, \PSP expands the potential applications of SSM, enabling its use in scenarios where complete and clean shape representations are unavailable, as well as facilitating specific clinical research endeavors such as multiclass and longitudinal/dynamic studies. By enhancing the accessibility and versatility of SSM, \PSP has the potential to advance clinical knowledge and improve patient care.

\section*{Acknowledgments}
The authors thank the National Institutes of Health for supporting this work under grant numbers NIBIB-U24EB029011, NIAMS-R01AR076120, NHLBI-R01HL135568, and NIBIB-R01EB016701.
The content is solely the responsibility of the authors and does not necessarily represent the official views of the National Institutes of Health. The authors also thank the ShapeWorks team, the University of Utah Division of Cardiovascular Medicine for providing the left atrium
MRI scans and segmentations from the Atrial Fibrillation projects, as well as the
Orthopaedic Research Laboratory (Andrew Anderson, PhD) at the University
of Utah for providing femur CT scans and corresponding segmentations.

{
\appendix

\section{Point Cloud Distance Metrics}
\label{app:distance}

A number of permutation-invariant distance metrics have been proposed for point cloud evaluation. Chamfer distance (CD) (\eqref{eq:CD}) is a popular choice because it does not require the point clouds to have the same number of points. Earth Mover's Distance (EMD) has been shown to outperform CD on learning representation tasks, but requires point clouds to have equal size and has significantly higher computational cost. 
Sliced Wasserstein distance (\textbf{SWD}) \cite{nguyen2021swd} is a more computationally efficient metric (based on projecting the points in point clouds onto a line) and has been shown to match EMD performance. 
Alternatively, Cauchy-Schwartz (\textbf{CS}) divergence, which calculates the divergence between two probability distributions, has been utilized \cite{he2023learning}. By representing point clouds as probability density functions and seeking to minimize divergence between them, this metric is more robust to outliers. However, it is also symmetric and requires the point clouds to be the same size. 

\begin{figure}[!h]
    \centering
    \includegraphics[width=.48\textwidth]{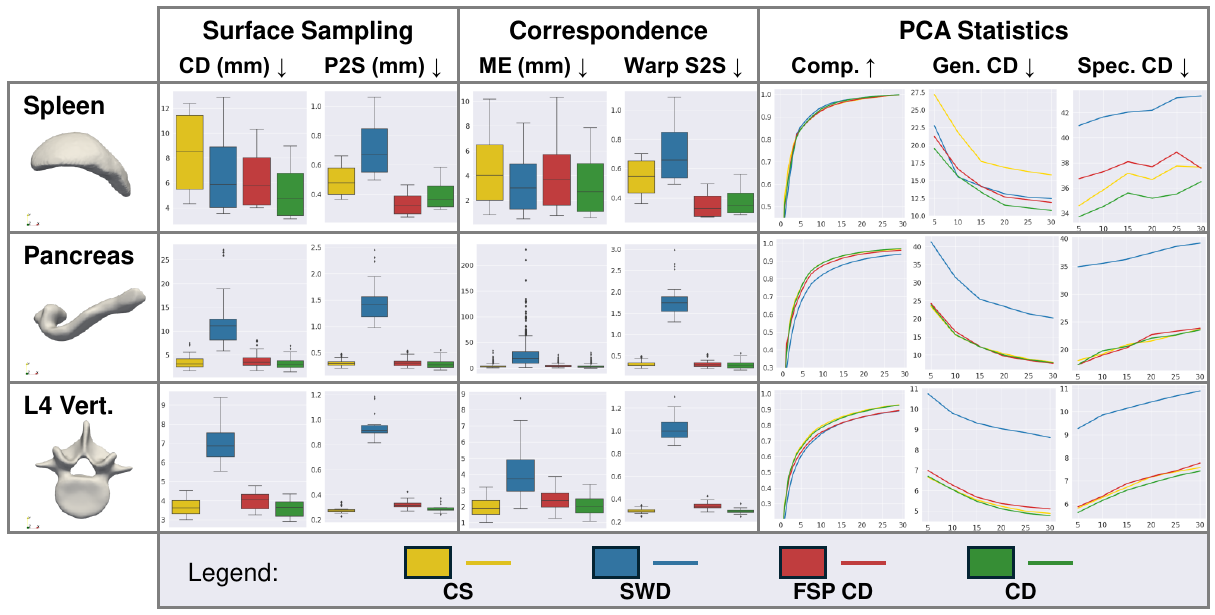}%
    \caption{Results with different loss distance metrics are reported across three datasets. Boxplots show the error distribution across test sets for each model in mm. Compactness plots show the cumulative variance ratio over the number of PCA modes. Generalization and specificity plots show the CD over the number of PCA modes.}
    \label{fig:distance}
\end{figure}

To empirically justify our choice of CD for the distance metric, we compare against SWD and CS across three datasets: spleen, pancreas, and L4 vertebrae. The loss was computed between $\sS$ and $\sY$ for each distance metric. As SWD and CS require point clouds to be the same size, a downsampled version of $\sS$ is created via furthest point sampling (FPS). For a fair comparison, we also compute CD using this downsampled version of $\S$, this metric is denoted \textbf{FPS CD}. The results are provided in \figref{fig:distance}.
The CD loss performs best across metrics, supporting our choice in the Point2SSM loss and reinforcing the robustness provided by requiring the predicted points to match a more dense point cloud.

\section{Correspondence Ablation Experiment}
\label{app:k_ablation}

The choice of $K$, or how many neighbors are used in graph construction for the DGCNN encoder, impacts the expressivity of the network. Here, we demonstrate this effect experimentally by training with different values of $K$, keeping $L$ fixed at 16 to limit the capacity of the feature representation. We perform this experiment using two loss functions: the self-supervised term used in Point2SSM $CD(\sS, sY)$, and an input reconstruction loss $CD(\sX, \sY)$. The results are provided in \figref{fig:k_ablation}.
With $\mathrm{CD}(\sX, \sY)$ loss, the best results are achieved with $k$ values between 8 and 64. With too few neighbors or too many, performance on variance metrics suffers. For example, with $K=1$ the CD is good, but compactness is poor. Whereas with $K=256$, CD is poor. While the reconstruction loss, $CD(\sX, \sY)$ is sensitive to the choice of $K$, the self-supervised loss $CD(\sS,\sY)$ is robust to the choice of $K$. 

\begin{figure}[!h]
    \centering
    \includegraphics[width=.48\textwidth]{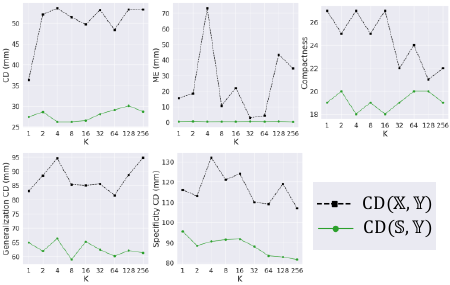}%
    \caption{Impact of $K$ hyperparameter values with self supervised loss $CD(\sS, \sY)$ and reconstruction loss $CD(\sX,\sY)$.}
    \label{fig:k_ablation}
\end{figure}

\section{Alignment Comparison}
\label{app:alignment}

Point2SSM assumed input point clouds are prealigned, whereas Point2SSM++ allows for misaligned input. To demonstrate this ability, we trained both on the same prealigned pancreas and spleen data. We then tested both models on three versions of the test datasets:
\begin{enumerate}
    \item Prealigned - test data aligned to match the prealigned training data
    \item Original - test data in the original coordinate system it was acquired from
    \item Misaligned - test data with additional added misalignment in the form of random translations and rotations.
\end{enumerate}

\begin{figure}[!h]
    \centering
    \includegraphics[width=.48\textwidth]{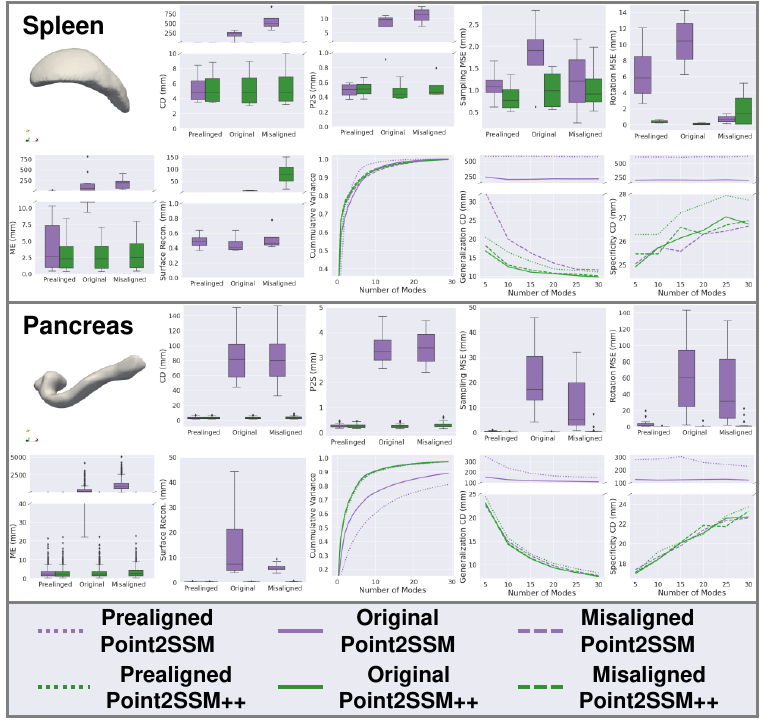}%
    \caption{\PS and \PSP results on prealigned, original, and misaligned data.}
    \label{fig:alignment}
\end{figure}

The results are provided in \figref{fig:alignment}. 
Point2SSM and Point2SSM++ achieve similar accuracy on the test dataset that is aligned to match the training data. However, the accuracy of Point2SSM significantly decreases when the test datasets are not prealigned. Point2SSM++ achieves similar accuracy on all three versions of the test datasets for both the spleen and pancreas. These results demonstrate the superiority of \PSP in handling misaligned input.

\section{Single Anatomy Experiments Statistical Significance}
\label{app:single_pvalues}

\begin{table}[!h]
\caption{Paired t-test p-values for the difference in error metrics between \PSP and each comparison method on single anatomy datasets. Values greater than 0.05, indicating non-significant differences at the 5$\%$ significance level, are highlighted in red.\label{tab:pvalues}}
\centering
\resizebox{.48\textwidth}{!}{%
\begin{tabular}{|l|l|llllll|}\hline
 Anatomy & Metric & DG-AE vs & ISR vs & DPC vs & SC3K vs & SC3K++ vs & Point2SSM vs \\ 
 & & \PSP & \PSP & \PSP & \PSP & \PSP & \PSP \\ \hline
 
\multirow{6}{*}{Spleen} 
& CD & 2.53e-02 & 2.79e-03 & {\color[HTML]{CB0000} 5.25e-02 } & 6.69e-03 & {\color[HTML]{CB0000} 1.88e-01 } & {\color[HTML]{CB0000} 5.92e-02 } \\
& P2S & 4.34e-03 & 6.29e-04 & {\color[HTML]{CB0000} 2.53e-01 } & 4.32e-03 & {\color[HTML]{CB0000} 2.51e-01 } & {\color[HTML]{CB0000} 5.43e-01 } \\
& Sampling MSE & {\color[HTML]{CB0000} 1.32e-01 } & 6.12e-03 & 2.46e-02 & {\color[HTML]{CB0000} 3.85e-01 } & {\color[HTML]{CB0000} 1.38e-01 } & {\color[HTML]{CB0000} 1.07e-01 } \\
& Rotation MSE & {\color[HTML]{CB0000} 6.10e-02 } & 2.78e-02 & {\color[HTML]{CB0000} 5.12e-02 } & {\color[HTML]{CB0000} 1.69e-01 } & {\color[HTML]{CB0000} 9.06e-01 } & 2.20e-03 \\
& ME & 1.76e-03 & 5.45e-04 & 2.68e-04 & 3.08e-03 & 6.66e-04 & 2.11e-03 \\
& Warp S2S & 5.01e-03 & 3.28e-04 & {\color[HTML]{CB0000} 1.12e-01 } & 4.03e-04 & {\color[HTML]{CB0000} 2.96e-01 } & {\color[HTML]{CB0000} 5.69e-01 } \\
\hline

\multirow{6}{*}{Pancreas} 
& CD & 3.86e-08 & 1.55e-05 & 4.39e-03 & 6.33e-13 & 6.68e-05 & 4.95e-05 \\
& P2S & 1.10e-13 & 2.96e-11 & 7.57e-06 & 8.47e-19 & 8.46e-12 & 3.09e-10 \\
& Sampling MSE & 1.60e-07 & 8.35e-08 & 3.01e-04 & 1.13e-07 & 2.09e-12 & 8.44e-08 \\
& Rotation MSE & 7.14e-09 & 3.07e-04 & 5.56e-06 & 2.71e-03 & 3.76e-02 & 9.54e-05 \\
& ME & 5.21e-119 & 1.09e-18 & 1.32e-76 & 1.00e-64 & 1.35e-100 & 2.16e-109 \\
& Warp S2S & 3.68e-13 & 1.87e-08 & 7.56e-06 & 4.86e-20 & 1.46e-11 & 2.49e-09 \\
\hline

\multirow{6}{*}{Liver} 
& CD & 4.59e-23 & 6.32e-55 & 2.98e-49 & 4.40e-28 & 2.78e-57 & {\color[HTML]{CB0000} 9.49e-01 } \\
& P2S & 1.32e-41 & 4.73e-77 & 1.02e-63 & 4.54e-20 & 1.72e-63 & 1.85e-06 \\
& Sampling MSE & 2.61e-03 & 8.04e-71 & 2.00e-43 & 3.44e-59 & 1.40e-73 & 1.70e-26 \\
& Rotation MSE & 1.25e-14 & 2.70e-11 & 2.72e-15 & 5.34e-12 & 4.14e-04 & 2.32e-17 \\
& ME & 0.00e+00 & 0.00e+00 & 0.00e+00 & 0.00e+00 & 0.00e+00 & 1.53e-02 \\
& Warp S2S & 5.90e-41 & 8.56e-81 & 1.19e-66 & 6.43e-29 & 3.01e-66 & 2.10e-02 \\
\hline

\multirow{6}{*}{Femur} 
& CD & 1.92e-02 & 8.26e-04 & 5.29e-04 & 2.25e-03 & 7.15e-04 & 1.93e-02 \\
& P2S & 5.77e-03 & 8.73e-05 & 1.35e-04 & 2.37e-03 & 3.60e-04 & 3.16e-05 \\
& Sampling MSE & {\color[HTML]{CB0000} 5.49e-01 } & 3.21e-03 & 6.12e-05 & 1.03e-03 & 7.82e-03 & 5.03e-03 \\
& Rotation MSE & 2.21e-02 & 2.18e-02 & 7.92e-04 & {\color[HTML]{CB0000} 1.18e-01 } & {\color[HTML]{CB0000} 1.71e-01 } & 6.92e-03 \\
& ME & 5.92e-03 & 2.23e-12 & 1.48e-11 & 4.62e-11 & 1.90e-13 & 3.32e-11 \\
& Warp S2S & 5.75e-03 & 6.07e-05 & 3.80e-05 & 1.12e-03 & 5.15e-04 & 1.25e-05 \\
\hline

\multirow{6}{*}{L4 Vert.}
& CD & 4.90e-08 & 4.40e-12 & 9.57e-11 & 2.14e-15 & 2.23e-13 & 2.28e-06 \\
& P2S & 1.30e-11 & 1.06e-14 & 6.28e-15 & 5.30e-16 & 3.10e-15 & 1.47e-04 \\
& Sampling MSE & 1.43e-03 & 3.08e-13 & 9.00e-11 & 8.49e-05 & 5.78e-14 & 1.89e-10 \\
& Rotation MSE & 2.47e-02 & 6.55e-03 & 8.05e-04 & 4.31e-04 & 3.40e-02 & 2.24e-02 \\
& ME & 3.07e-65 & 1.62e-136 & 2.89e-149 & 4.48e-52 & 1.05e-169 & 1.94e-60 \\
& Warp S2S & 1.84e-11 & 2.26e-16 & 6.99e-16 & 1.52e-15 & 1.09e-15 & 3.15e-05 \\
\hline

\multirow{6}{*}{Cranium} 
& CD & {\color[HTML]{CB0000} 3.17e-01 } & 9.66e-31 & 2.22e-17 & 1.55e-36 & 1.43e-29 & 5.93e-03 \\
& P2S & {\color[HTML]{CB0000} 7.91e-02 } & 9.42e-44 & 6.97e-25 & 1.13e-47 & 5.86e-37 & {\color[HTML]{CB0000} 1.52e-01 } \\
& Sampling MSE & 6.35e-15 & 1.21e-33 & 4.57e-29 & 3.98e-25 & 1.01e-29 & 9.07e-16 \\
& Rotation MSE & 4.51e-09 & 3.23e-06 & 7.20e-09 & {\color[HTML]{CB0000} 1.22e-01 } & 6.85e-03 & 1.83e-08 \\
& ME & 0.00e+00 & 0.00e+00 & 0.00e+00 & 0.00e+00 & 0.00e+00 & 2.69e-05 \\
& Warp S2S & {\color[HTML]{CB0000} 5.26e-02 } & 1.72e-42 & 4.62e-16 & 5.06e-51 & 1.33e-37 & {\color[HTML]{CB0000} 4.69e-01 } \\
\hline

\end{tabular}%
}
\end{table}

\revise{To quantitatively assess whether the differences in error metrics shown in Figure \ref{fig:benchmark} are statistically significant, we perform paired t-tests. For each metric, the errors produced by \PSP and each comparison method are treated as paired observations, and the p-value is computed for the null hypothesis that the mean difference between them is zero. The resulting p-values, reported in Table \ref{tab:pvalues}, indicate the likelihood that the observed performance differences occurred by chance.}
\revise{
The majority of p-values are below 0.05, indicating that \PSP achieves a statistically significant reduction in error. On each dataset, \PSP shows a significant improvement over every comparison method for at least two of the six evaluated metrics. This analysis provides strong statistical support for the performance gains achieved by the proposed method relative to existing approaches.}

\section{Additional Modes of Variation}
\label{app:modes}

The modes of variation for all comparison models on the single anatomy experiments are provided in \figref{fig:extra_modes}. The reconstructed meshes resulting from the \SC model have the most artifacts, indicating miscorrespondence. Some artifacts can also be seen in the \ISR and \DPC mesh reconstructions. The modes of variation are generally similar to those captured by the \PSM and \PSP models (shown in \figref{fig:benchmark_modes}). Overall, \PSP provides the smoothest and most interpretable modes of variation. 

\begin{figure*}[!ht]
    \centering
    \includegraphics[width=\textwidth]{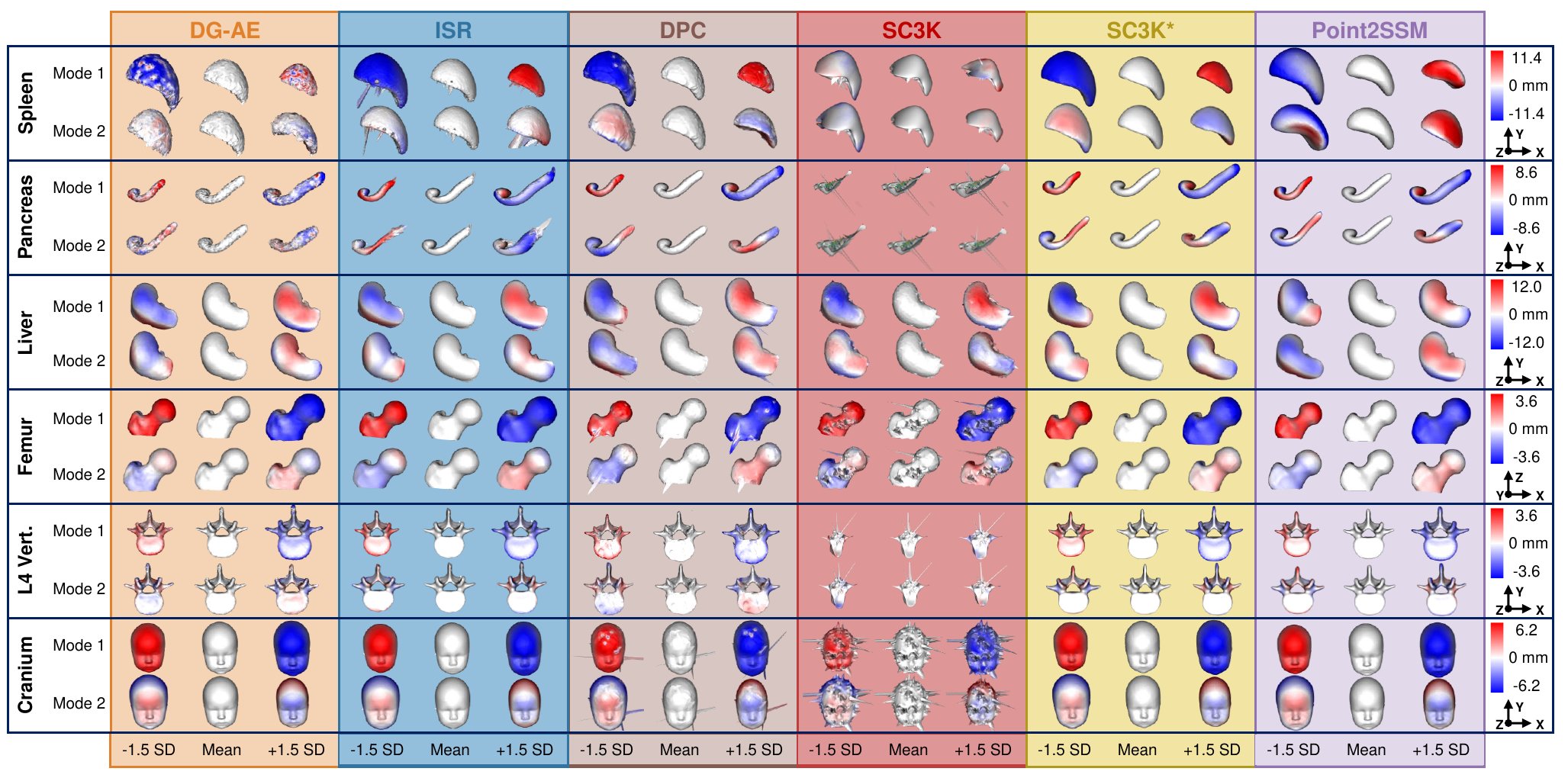}%
    \caption{Primary and secondary modes of population variation captured by comparison methods across each dataset. The mean and shapes at $\pm 1.5$ standard deviation are shown with color denoting the distance to the mean. Blue indicates the surface is outside the mean, and red indicates the surface is inside the mean. Separate color scales and orientation markers are provided for each dataset.}
    \label{fig:extra_modes}
\end{figure*}

\section{Comparison with FlowSSM}
\label{app:flowssm}

\edit{Flow-based statistical shape models (SSMs), such as FlowSSM \cite{ludke2022landmark}, represent a state-of-the-art family of deformation-based approaches that construct population shape models by learning diffeomorphic mappings from a template or mean shape to individual instances. These methods have demonstrated strong performance in settings where a canonical surface representation and dense registration are available.}

\edit{Point2SSM++, in contrast, is designed to operate directly on unordered point clouds without requiring a template surface, segmentation, or predefined parameterization. Due to these fundamentally different assumptions, a direct comparison between Point2SSM++ and FlowSSM is not fully aligned. Nevertheless, to provide additional context on Point2SSM++ performance, we include supplementary comparison experiments in this appendix. These experiments are intended to offer qualitative and quantitative insight rather than to serve as a definitive benchmark.}

\edit{To enable comparison with FlowSSM, additional preprocessing steps were required that fall outside the native Point2SSM++ pipeline. Specifically, a template surface had to be defined and used as the reference shape for FlowSSM training, following the procedure described in \cite{ludke2022landmark}. We selected the shape in the cohort that was closest to the mean obtained from the \PSM approach. In this way, the most representative shape was used as the template, as FlowSSM is sensitive to template selection \cite{iyer2023mesh2ssm}.}

\edit{FlowSSM was trained using the official implementation and default hyperparameters, with adjustments made only where necessary to accommodate the dataset used in this study. Point2SSM++ was trained directly on the original point clouds without access to surface information or a template shape.
All experiments were conducted on the same datasets described in \label{sec:datasets} and evaluated on an identical train/test split and metrics to ensure comparability.}

\begin{figure*}[!ht]
    \centering
    \includegraphics[width=\textwidth]{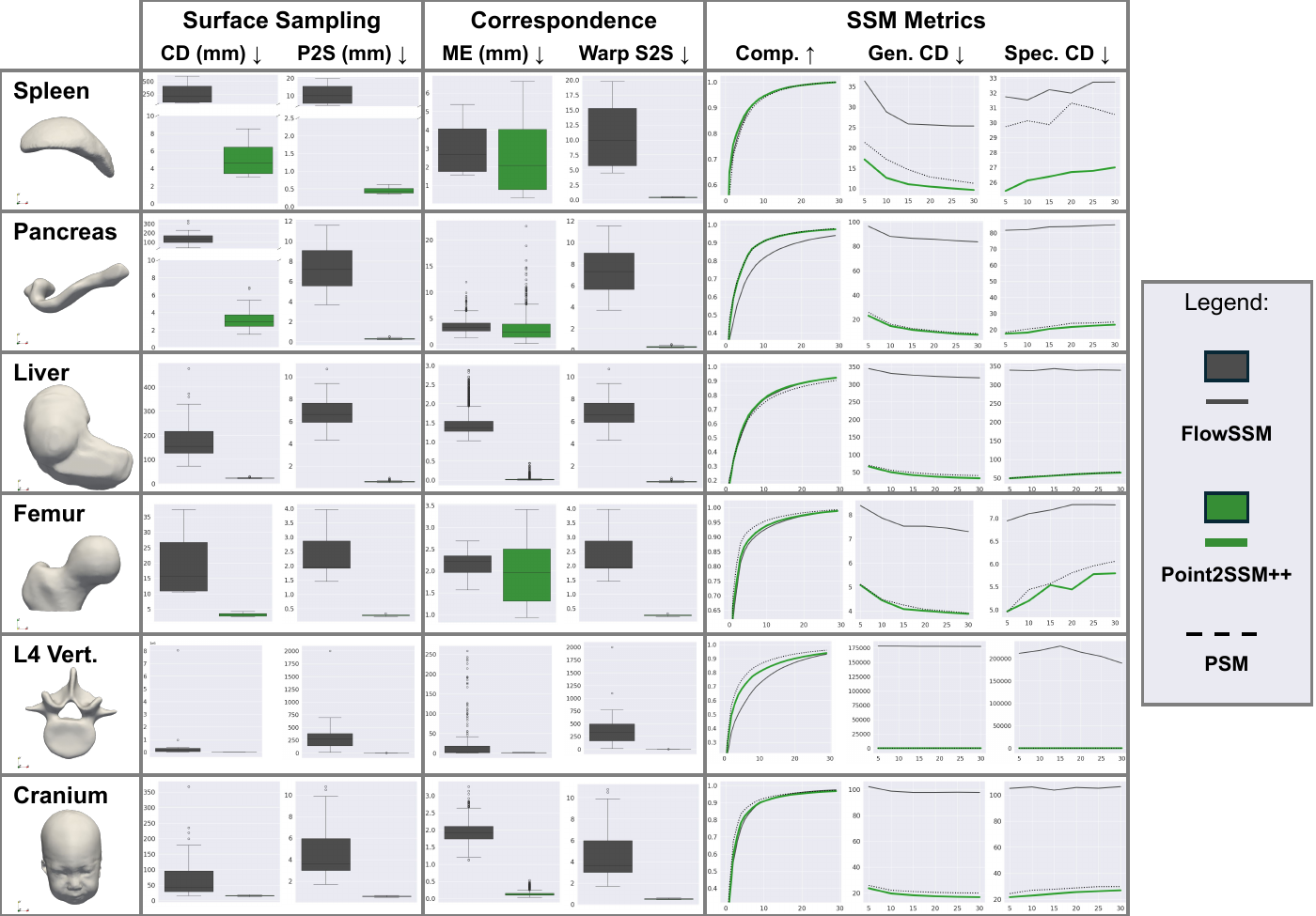}%
    \caption{Comparison of FlowSSM \cite{ludke2022landmark} and Point2SSM++ on single anatomy experiments. Boxplots show the error distribution across test sets for each model in mm. Compactness plots show the cumulative variance ratio over the number of PCA modes. Generalization and specificity plots show the CD over the number of PCA modes. A maximum of 30 modes is displayed which captures at least $99\%$ of the total variation for all datasets. Lower values are better for all metrics except PCA compactness.}
    \label{fig:flowssm}
\end{figure*}

\edit{Quantitative comparison results are summarized in Figure \ref{fig:flowssm}. All metrics described in Section \ref{sec:metrics} are included with the exception of the consistency metrics, given that FlowSSM cannot operate on different input samplings of the same shape. Point2SSM++ outperforms FlowSSM on all metrics across datasets despite operating without access to surface topology or a predefined mean shape. These results suggest that while FlowSSM is well suited for template-based pipelines, Point2SSM++ offers a viable alternative in scenarios where such preprocessing is unavailable or undesirable.}

\section{Multi-Anatomy Experiments Statistical Significance}
\label{app:multi_pvalues}

\revise{This section provides statistical significance analysis for the multi-anatomy experiments presented in Section \ref{sec:multianatomy}, similar to the analysis performed for single anatomy experiments in \ref{app:single_pvalues}.} 

\begin{table}[!h]
\caption{Paired t-test p-values for the difference in error metrics between single and multiclass models on the multi-organ dataset. Values greater than 0.05, indicating non-significant differences at the 5$\%$ significance level, are highlighted in red.\label{tab:multi_pvalues}}
\centering
\resizebox{.48\textwidth}{!}{%
\begin{tabular}{|l|l|cc|}\hline
 Anatomy & Metric & Single-Class \PSP  & Single-Class \PSP  \\
 & & vs & vs \\
 & & Multiclass \PSP & Multiclass \PSP, Classifier \\ \hline
 
\multirow{6}{*}{Spleen} 
& CD & {\color[HTML]{CB0000} 5.63e-02 } & 2.56e-02 \\
& P2S & 1.57e-02 & 1.17e-02 \\
& Samp. MSE & {\color[HTML]{CB0000} 5.18e-02 } & 4.62e-02 \\
& Rot. MSE & {\color[HTML]{CB0000} 2.58e-01 } & {\color[HTML]{CB0000} 1.37e-01 } \\
& ME & 2.54e-04 & 2.06e-04 \\
& Warp S2S & 2.35e-02 & 1.77e-02 \\
\hline

\multirow{6}{*}{Pancreas} 
& CD & 4.54e-02 & {\color[HTML]{CB0000} 2.50e-01 } \\
& P2S & 2.71e-06 & 6.04e-07 \\
& Samp. MSE & {\color[HTML]{CB0000} 9.64e-01 } & 7.80e-03 \\
& Rot. MSE & {\color[HTML]{CB0000} 2.57e-01 } & {\color[HTML]{CB0000} 4.25e-01 } \\
& ME & 4.60e-09 & 5.67e-09 \\
& Warp S2S & 2.89e-07 & 1.56e-08 \\
\hline

\multirow{6}{*}{Liver} 
& CD & 2.08e-17 & 5.39e-23 \\
& P2S & 1.80e-34 & 1.05e-39 \\
& Samp. MSE & 2.65e-07 & 1.16e-08 \\
& Rot. MSE & {\color[HTML]{CB0000} 1.53e-01 } & 3.88e-02 \\
& ME & 2.08e-52 & 3.75e-85 \\
& Warp S2S & 1.39e-32 & 1.70e-33 \\
\hline

\end{tabular}%
}
\end{table}

\begin{table}[!h]
\caption{Paired t-test p-values for the difference in error metrics between single and multiclass models on the multiclass vertebrae dataset. Values greater than 0.05, indicating non-significant differences at the 5$\%$ significance level, are highlighted in red.\label{tab:vert_pvalues}}
\centering
\resizebox{.48\textwidth}{!}{%
\begin{tabular}{|l|l|cc|}\hline
 Vertebrae & Metric & Single-Class \PSP vs & Single-Class \PSP vs \\
 Type & & Multiclass \PSP & Multiclass \PSP, Classifier \\ \hline
\multirow{4}{*}{Cervical} 
& CD & 1.27e-114 & 2.14e-107 \\
& Sampling MSE & 1.04e-80 & 8.58e-70 \\
& Rotation MSE &  {\color[HTML]{CB0000} 1.34e-01 } & 7.02e-06 \\
& ME & 1.14e-03 & 9.55e-05 \\
\hline

\multirow{4}{*}{Thoracic} 
& CD & 3.28e-44 & 4.63e-41 \\
& Sampling MSE & 6.04e-92 & 3.93e-68 \\
& Rotation MSE &  {\color[HTML]{CB0000} 7.40e-01 } & 1.29e-02 \\
& ME & 4.65e-79 & 1.37e-72 \\
\hline

\multirow{4}{*}{Lumbar} 
& CD & 2.52e-40 & 6.13e-37 \\
& Sampling MSE & 4.67e-67 & 1.13e-20 \\
& Rotation MSE & 6.08e-10 & 1.27e-03 \\
& ME & 6.04e-91 & 6.31e-54 \\
\hline

\end{tabular}%
}
\end{table}

\revise{P-values for paired t-tests between the single and multiclass models are reported in Table \ref{tab:multi_pvalues} for the multi-organ experiment (Figure \ref{fig:multi_organ}) and Table \ref{tab:vert_pvalues} for the multiclass vertebrae experiment (Figure \ref{fig:multi_vert}). For brevity, observations are aggregated across the three vertebrae types in Table \ref{fig:multi_vert}: cervical (C1-C7), thoracic (T1-T12), and lumbar (L1-L5). The p-values suggest the error improvement of the multiclass models are statistically significant in most cases, especially for the multiclass \PSP model with auxiliary classifier.}


\section{Training Time Comparison}
\label{app:runtime}

\revise{Table \ref{tab:runtime} compares the average time required to train the deep learning approaches. The values are an average of the time taken to train the models on a 4x TITAN V GPU on each of the datasets described in Section \ref{sec:datasets}, weighted by the size of the dataset. This provides the average train time required per shape in seconds. \PSP averaged 22.2 seconds per shape; thus, training \PSP on a cohort of 100 shapes of average complexity may take around 37 minutes. The \PSP model does not require significantly more training time than any of the comparison methods, including the base \PS model.}
\begin{table}[!h]
\caption{ Average training times are reported for various models in seconds per shape in the training cohort. \label{tab:runtime}}
\centering
\resizebox{.3\textwidth}{!}{%
\begin{tabular}{|l|c|}\hline
Model & Avg. train time (sec/shape) \\ \hline
\DAE & 21.4 \\
\ISR & 14.3 \\
\DPC & 11.5 \\
\SC &  15.3 \\
\SCP & 26.0 \\
\PS &  18.5 \\
\PSP & 21.4\\ \hline
\end{tabular}%
}
\end{table}


 \bibliographystyle{elsarticle-num-names} 
 \bibliography{references}

\end{document}